\pdfoutput=1
\documentclass{article}
\usepackage[final]{colm2026_conference}

\usepackage{microtype}
\usepackage{hyperref}
\usepackage{url}
\usepackage{booktabs}
\usepackage{graphicx}
\usepackage{tabularx}
\usepackage{array}
\usepackage{amsmath}

\usepackage{placeins}
\usepackage{etoolbox}

\definecolor{darkblue}{rgb}{0, 0, 0.5}
\hypersetup{colorlinks=true, citecolor=darkblue, linkcolor=darkblue, urlcolor=darkblue,
  pdftitle={ComponentBench: Diagnosing Component-Level Failures in Computer-Use Agents},
  pdfauthor={Tianchen Guan, Xinlei Lin, Royce Cheng-Yue, Xiangjun Wang, Shuyan Zhou}}

\title{ComponentBench: Diagnosing Component-Level Failures in Computer-Use Agents}

\author{%
Tianchen Guan$^{1}$ \qquad Xinlei Lin$^{2}$ \qquad Royce Cheng-Yue$^{2}$ \\
\textbf{Xiangjun Wang}$^{2}$ \qquad \textbf{Shuyan Zhou}$^{1}$ \\
\texttt{\{tianchen.guan, shuyan.zhou\}@duke.edu}\\
$^{1}$Duke University \qquad $^{2}$Amazon AGI SF Lab
}

\begin{document}
\maketitle

\begin{abstract}
Current evaluation of computer-use agents is split between long-horizon workflow benchmarks and atomic GUI-grounding tests. This leaves an under-instrumented middle layer: realistic component-centered interactions (e.g., toggle a button set) that are short enough to diagnose and rich enough to capture the burdens of modern interfaces. We present ComponentBench, a benchmark and diagnostic pipeline for component-level evaluation of computer-use agents on modern web UIs. ComponentBench is organized around a library-agnostic ontology of 97 canonical UI components instantiated as 2,910 programmatically verified tasks across widely used component libraries, paired with cleaned human reference trajectories that enable evaluation of both task success and interaction efficiency. Beyond task collection, we introduce a scalable pipeline for auditing realized structural difficulty after implementation and synthesizing structured failure analyses across tasks and component families. 
Evaluating seven models---GPT-5.4, Gemini~3~Flash, GPT-5.4~mini, GPT-5~mini, Gemini~3.1~Flash-Lite, Qwen3-VL-235B, and UI-TARS-1.5-7B---across four observation and action spaces, we show that these design choices critically impact performance. Within a single shared harness, changing only the observation and action space shifts task success by more than 30\% for the same model: GPT-5~mini falls from 83.1\% with accessibility-tree observations to 48.9\% with coordinate-only Pixel control.
Moreover, even the fastest configuration takes 3.7$\times$ as long as the matched human reference, and spatial manipulations that are trivial for humans continue to challenge current agents.
\end{abstract}

\section{Introduction}

Computer-use agents are moving from research prototypes toward user-facing systems that act on websites and software through the same interfaces people use. OpenAI's Operator \citep{openai2025operator} and Computer Use API \citep{openai2026computeruse}, along with Anthropic's computer-use tool \citep{anthropic2026computeruse}, make the rendered interface itself---typically screenshots plus mouse and keyboard actions---a first-class control surface. This makes full-visual evaluation increasingly central. As screenshot-native agents become stronger, the key question is no longer simply whether an agent can occasionally complete a browser task, but which rendered UI components still prevent reliable and efficient use in deployment.

Current evaluation paradigms still emphasize two extremes. Long-horizon benchmarks such as WebArena \citep{zhou2023webarena}, VisualWebArena \citep{koh2024visualwebarena}, OSWorld \citep{xie2024osworld}, WebVoyager \citep{he2024webvoyager}, and Online-Mind2Web \citep{xue2025illusion} measure end-to-end competence on realistic tasks, but make failure attribution difficult: when an agent misses a workflow, the root cause may be planning, state tracking, grounding, or one brittle interaction buried inside a larger task. At the other extreme, grounding benchmarks such as ScreenSpot \citep{cheng2024seeclick} and ScreenSpot-Pro \citep{li2025screenspotpro} isolate localization ability, but stop before the short multi-step interactions that modern widgets often require. Efforts between the two---MiniWoB++'s synthetic micro-environments \citep{liu2018workflow}, Mind2Web's offline real-site traces \citep{deng2023mind2web}, and the web-action taxonomies, archived GUI subtasks, and appearance variations of WebSuite, WARC-Bench, and OpenApps \citep{li2024websuite,srivastava2025warc,ullrich2025openapps}---remain partial for our purpose: to our knowledge, none are organized around a broad cross-library ontology of modern UI components with programmatic end-state verification, human reference traces, and post-render difficulty auditing.

\begin{figure*}[!t]
\centering
\includegraphics[width=\textwidth]{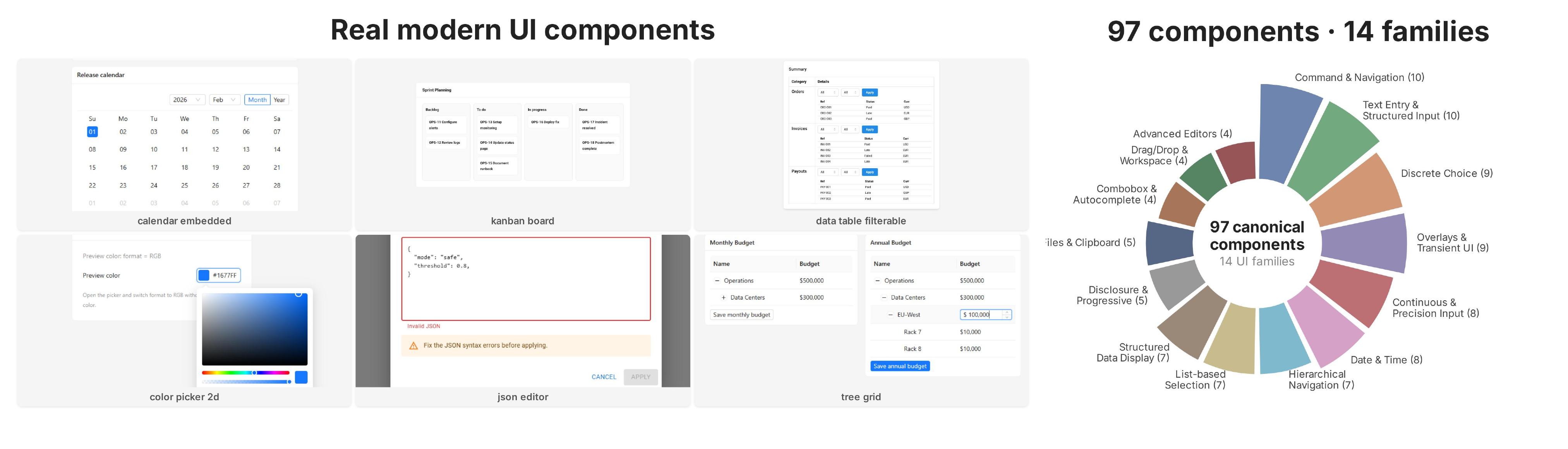}
\vspace{-2pt}
\caption{\textbf{ComponentBench} evaluates computer-use agents on 97 canonical UI component types organized into 14 families. \textbf{Left}: Tasks span diverse interaction types implemented across Ant Design, MUI, and Mantine. \textbf{Right}: The ontology covers the breadth of modern web UI interaction.}
\label{fig:hero}
\end{figure*}

This missing layer matters because modern web software is composed from recurring component primitives exposed at scale by production UI libraries. A long workflow can therefore fail not because the agent misunderstood the user's goal, but because it mishandled one date picker, multiselect, splitter, or drag target. Long tasks are only as reliable as the component interactions they contain: under a simple independence approximation, five critical interactions at 80\% reliability imply only about a 33\% end-to-end ceiling. And pass rate alone is insufficient. Recent work shows that even strong computer-use agents often take far more steps than humans, with major latency and cost implications for real deployment \citep{abhyankar2025osworldhuman}.

We introduce \textbf{ComponentBench},\footnote{Website: \url{https://componentbench.com}; code: \url{https://github.com/TianchenGuan/ComponentBench}; data and runs: \url{https://huggingface.co/datasets/TianchenGuan/ComponentBench}.} a benchmark and diagnostic pipeline for evaluating computer-use agents through component-centered tasks on modern web UIs. ComponentBench organizes evaluation around \textbf{97 canonical component types}, \textbf{14 interaction families}, and \textbf{24 task templates}, instantiated as \textbf{2,910} programmatically verified tasks primarily across Ant Design, MUI, and Mantine (thirty markdown-editor tasks use an external implementation; Appendix~\ref{app:task-diversity}). Each task is anchored to a single primary component, even when realistic carrier context is required. The benchmark evaluates agents under four observation and action spaces---AX-tree, Set-of-Marks, Pixel, and Browser-Use---and distinguishes intended difficulty from the difficulty the rendered UI actually presents through replay-based audits using human reference traces.

ComponentBench is also designed to evaluate efficiency, not only eventual completion. Because we collect cleaned human reference traces for all tasks, we can ask not just whether a component is solvable, but whether it is solved directly enough to be usable---critical for full-visual agents, where every extra step implies more latency, more token cost, and another opportunity to drift. To support faster stress-testing, we further derive \textbf{ComponentBench-Core}, a distilled \textbf{912}-task hard-only suite of newly generated tasks that concentrates on the unresolved regions of the full suite.

Our experiments on ComponentBench-Full across seven models and four observation/action spaces reveal four main findings. First, observation/action space can shift pass rates by more than 30\% within a single model, and the benefit of Set-of-Marks is model-dependent rather than universal. Second, efficiency remains a major deployment bottleneck: even the fastest configuration takes 3.7$\times$ as long as the matched human reference, and the strongest model solves many more tasks eventually than within the human step budget. Third, several spatial manipulation components that humans finish in 1--2 steps---including sliders, drag-and-drop lists, and splitters---remain below 60\% mean pass rate across all agents tested. Finally, difficulty is strongly conditioned on visual context, with substantially wider AX-tree--Pixel gaps under clutter and compact spacing. A trace-grounded failure taxonomy (Section~\ref{sec:failure-taxonomy}) ties these findings to concrete mechanisms. Together, these results show that component-level evaluation exposes failure modes largely invisible in both long-horizon task scores and single-step grounding benchmarks.

\section{Benchmark construction}
\label{sec:benchmark-construction}
\begin{figure*}[t]
\centering
\includegraphics[width=\textwidth]{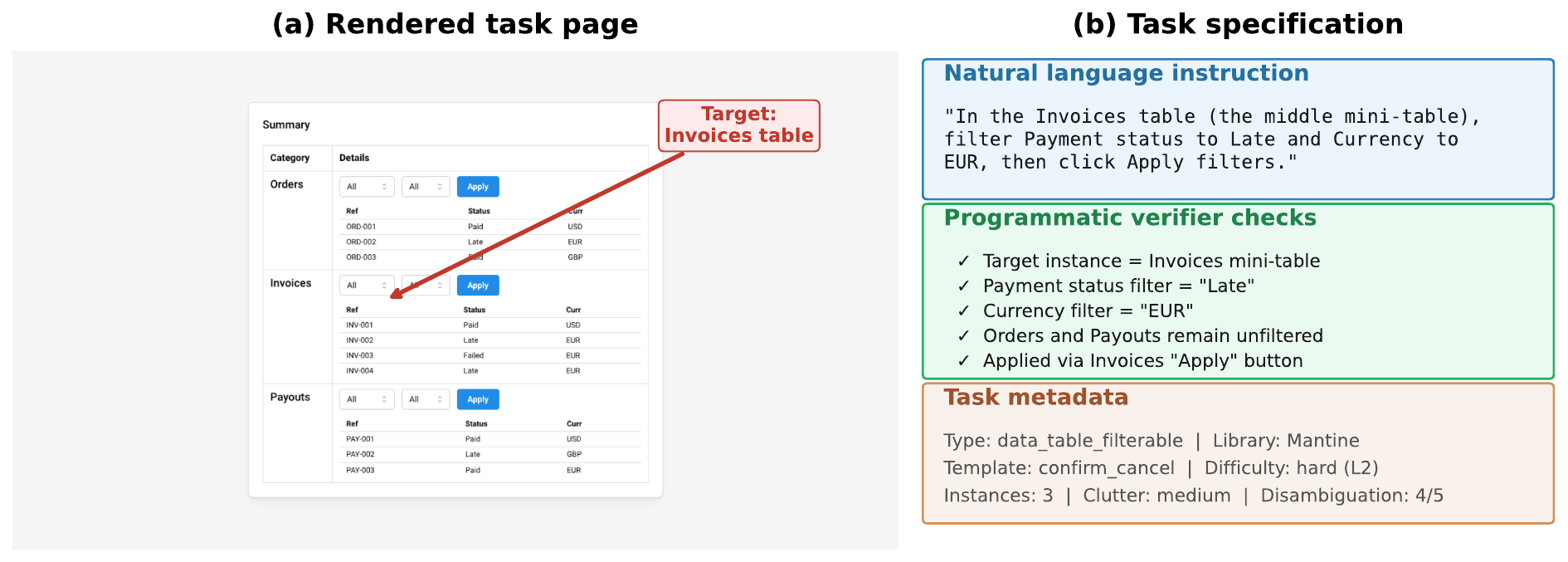}
\caption{A concrete ComponentBench task (\texttt{data\_table\_filterable-mantine-T10}). \textbf{(a)}~The agent sees a rendered page containing three visually similar mini-tables. \textbf{(b)}~The instruction, programmatic verifier, and structured metadata. The agent must disambiguate the correct table instance (Invoices), apply two filters, and commit via the instance-local Apply button.}
\label{fig:worked-example}
\end{figure*}

\paragraph{A worked example.}
Before introducing the benchmark schema, we start with a concrete task. Figure~\ref{fig:worked-example} shows \texttt{data\_table\_filterable-mantine-T10}. The page contains three visually similar mini-tables labeled \emph{Orders}, \emph{Invoices}, and \emph{Payouts}. The agent must operate only the \emph{Invoices} instance: set \emph{Payment status} to \emph{Late}, set \emph{Currency} to \emph{EUR}, and then click the local \emph{Apply} button. This single example already illustrates several recurring design choices in ComponentBench: each task targets one primary component type; the page may include surrounding \emph{carrier context} that adds realism without changing what is being tested; nearby instances can create disambiguation burden; and success is defined by a \emph{committed end state}, not by a draft selection.

\subsection{Component inventory and task specification}

To cover modern web interaction systematically, we build a structured inventory by combining the WAI-ARIA Authoring Practices Guide (APG), which documents common widget patterns and behaviors, with the production component inventories of major React UI libraries---Ant Design, MUI, Mantine, Fluent UI, Chakra UI, and Headless UI \citep{w3c_apg,antdesign_components,mui_components,mantine_components,fluentui_components,chakraui_components,headlessui_components}. The resulting \emph{component ontology} is a library-agnostic set of \textbf{97 canonical component types} (e.g., \emph{date picker}) grouped into \textbf{14 families} (e.g., \emph{Drag/Drop \& Workspace}). For implementation we select Ant Design, MUI, and Mantine for their comprehensive coverage and stylistic diversity, which also lets cross-library variation serve as a controlled factor in the analysis.

Each task evaluates \textbf{one primary component type}, even when the rendered page contains many other controls: in Figure~\ref{fig:worked-example}, a \emph{filterable data table}, with the enclosing summary table and neighboring mini-tables serving as carrier context.

Tasks are specified in YAML. Each specification includes a canonical type, an implementation source, a task template (a reusable action pattern such as \emph{open-and-select} or \emph{drag operation}), a scene context (eight controlled factors: theme, spacing, layout, placement, scale, instances, guidance, clutter), an intended difficulty block (seven conceptual axes: precision requirement, target acquisition, density/choice interference, depth/layering, feedback dynamics, semantic observability, disambiguation load), a success trigger, and negative cases. The full suite contains \textbf{24 canonical task templates} (plus one ad-hoc variant used by a single task) and \textbf{2,910 tasks}; Appendix~\ref{app:task-diversity} summarizes the realized diversity across all of these dimensions.

\subsection{Task generation, implementation, and human verification}
\label{sec:construction-human}

ComponentBench was built through a structured, LLM-assisted pipeline. First, GPT-5.2~Pro generated 30 task specifications per canonical type under the shared YAML schema, including task templates, scene factors, intended difficulty labels, success triggers, and negative cases. Second, Claude~Code implemented these specifications as real interactive Next.js pages rather than static mockups. Third, each implemented task was executed \emph{twice} by a human operator, recorded as low-level actions (clicks, drags, keyboard input, scrolls) with timestamps, cleaned---merging consecutive keystrokes and removing accidental resets---and the shorter pass was kept as the reference trajectory.

The human recordings play two roles. During construction, they are the decisive validity check that the task is solvable and faithful to the intended specification; tasks that were broken, miswired, or no longer matched their specification were corrected before inclusion. Later, the same cleaned traces serve as reference trajectories for efficiency analysis and replay-based difficulty auditing (a validation study with two additional annotators appears in Appendix~\ref{app:human-validation}). Across all 2,910 tasks, the cleaned traces have a mean of 2.7 normalized steps (median 2), with 97.8\% of tasks solvable in 10 or fewer steps and a mean completion time of 4.8 seconds.

\subsection{Programmatic verification}

Every task is paired with a \textbf{programmatic verifier} that checks the \emph{committed end state}. The benchmark therefore does not ask whether the agent briefly opened the right menu or drafted the right intermediate selection; it asks whether the underlying task predicate is satisfied after the relevant interaction has actually been committed. For some tasks, the live state is sufficient. For others, success requires an explicit local control such as \emph{Apply}, \emph{Save}, \emph{OK}, or \emph{Confirm}.

For the running example in Figure~\ref{fig:worked-example}, success requires that the \emph{Invoices} mini-table---and not the neighboring \emph{Orders} or \emph{Payouts} tables---has \emph{Payment status = Late} and \emph{Currency = EUR}, and that these selections have been committed via the instance-local \emph{Apply} button. Each YAML task also enumerates negative cases so that nearby but incorrect states do not count as success.

Concretely, the YAML specifies a canonical success predicate, and the page implementation provides a JavaScript checker that evaluates this predicate against the live component state. At the environment level, termination is intentionally simple and deterministic: success is signaled by revealing a benchmark banner (\texttt{\#cb-success-banner}). This gives all observation/action spaces the same termination condition while leaving the checker logic task-specific underneath. Importantly, verifier state is isolated from agent observation. Target values and success predicates reside in React component closures and are never exposed as DOM attributes, accessible-tree labels, or page text; the success banner appears only \emph{after} the correct state is already reached and therefore cannot be used to shortcut the task. In benchmark mode, a \texttt{MutationObserver} additionally strips all test-only DOM attributes (\texttt{data-testid}, \texttt{data-cy}, etc.) from every element in real time, and a CI-ready scanner verifies that no such attributes survive across all 2,910 pages.

\subsection{Observation and action spaces}
\label{sec:obs-regimes}

A core goal of ComponentBench is to evaluate the \emph{same underlying task} under different observation and action spaces. The benchmark therefore supports four regimes.

\textbf{AX-tree.} The agent receives a screenshot plus accessibility-tree text and acts through element IDs.

\textbf{Set-of-Marks (SoM).} The agent receives a screenshot with numbered overlays on interactable elements and also acts through element IDs.

\textbf{Pixel.} The agent receives only the screenshot and must act through coordinates.

\textbf{Browser-Use.} The same tasks are executed through the separate \texttt{browser-use} framework, which provides screenshot interaction together with a richer tool surface including serialized DOM information and grounded elements \citep{browseruse2024}. A Browser-Use ``step'' is a \emph{model turn} that can execute up to 4 primitive actions, effectively giving it up to 80 actions within the 20-step budget, compared to exactly 1 action per step in the other three modes.

The first three regimes share the same BrowserGym-based harness and differ only in what the agent sees and how it refers to targets \citep{dechezelles2024browsergym}. Browser-Use is therefore not a strictly like-for-like fourth mode but a \emph{tool-rich reference regime}---a distinction that lets us separate gains from stronger underlying models from gains that come from exposing additional structure and control affordances.

\subsection{Intended difficulty and realized difficulty}

ComponentBench distinguishes between the difficulty a task is \emph{designed} to have and the difficulty the rendered page \emph{actually} presents. \textbf{Intended difficulty} is assigned during task generation: the seven conceptual axes and coarse tier labels record why we expect the task to be easy or hard (for the running example, mainly the disambiguation among similar instances). Because implementation can change difficulty through spacing, overlay structure, or clutter, we also measure \textbf{realized difficulty} by replaying the cleaned human trace in Playwright \citep{playwright_eval} and extracting a \textbf{24-feature} structural vector---including target size, spacing, overlay depth, option density, and metadata-derived properties such as control continuity and reset cost. The human trace serves as a \emph{traversal oracle}, reaching hidden states (popovers, nested panels, internal scroll regions) that a static snapshot would miss. Features are aggregated into the same seven axes via weighted averaging with frozen cutpoints. In short, intended difficulty says what we meant to build; realized difficulty says what the implemented task actually is.

\subsection{Diagnostic pipeline and Core distillation}
\label{sec:full-to-core}

Beyond the executable benchmark itself, ComponentBench includes a three-stage diagnostic pipeline that turns tasks, traces, and model runs into reusable evidence: \textbf{Layer~1} deterministically assembles per-task packets joining the YAML specification, task metadata, realized-difficulty outputs, human reference traces, and selected agent evidence; \textbf{Layer~2} produces a structured task-level observation for each task; \textbf{Layer~3} aggregates those records into one component-level report per canonical type. This pipeline is what makes the benchmark diagnostic rather than purely leaderboard-oriented.

The full suite is intentionally coverage-oriented. Once broad evidence exists, we use realized-difficulty audits, Layer~2/3 reports, and scene-factor analyses to distill a sharper pressure test, \textbf{ComponentBench-Core}: \textbf{19 generation units}, \textbf{45 canonical components}, and \textbf{912 hard-only tasks}, preserving the same pipeline while concentrating benchmark mass on unresolved interaction families.

\section{Experiments}

This section focuses on ComponentBench-Full (v1), the broad coverage suite, because it provides the cleanest setting for comparing models across observation and action regimes. All tasks are run with a maximum budget of 20 agent steps. Our goal is not only to rank models, but to isolate how much of component-level performance comes from the underlying model versus the interface exposed to it.

\subsection{Models and regimes}

We evaluate seven models on v1 in up to four observation/action spaces. \textbf{Gemini~3~Flash} \citep{google2025gemini3flash} and \textbf{Gemini~3.1~Flash-Lite} \citep{google2026gemini31flashlite} are Google's multimodal models at different capability tiers. \textbf{GPT-5.4} \citep{openai2026gpt54}, \textbf{GPT-5.4~mini} \citep{openai2026gpt54mini}, and \textbf{GPT-5~mini} \citep{openai2025gpt5} are OpenAI models spanning full-scale and compact tiers. \textbf{Qwen3-VL-235B} \citep{wang2025qwen3vl} is an open-weight vision-language model. \textbf{UI-TARS-1.5-7B} \citep{qin2025uitars} is a 7B open-weight model trained for GUI interaction, evaluated through its own native pixel-only harness rather than the shared BrowserGym harness.

The four observation/action spaces are defined in Section~\ref{sec:obs-regimes}: AX-tree, SoM, and Pixel share the same benchmark harness and differ only in what the agent sees and how it refers to targets, while Browser-Use is the separate tool-rich reference regime \citep{browseruse2024}. The five Gemini and GPT models and Qwen3-VL-235B are evaluated in all four spaces; UI-TARS-1.5-7B in its native pixel-only mode.

\subsection{Main results}
\label{sec:main-results}

\begin{table}[t]
\centering
\small
\begin{tabular}{@{}lcccc@{}}
\toprule
\textbf{Model} & \textbf{Browser-Use} & \textbf{AX-tree} & \textbf{SoM} & \textbf{Pixel} \\
\midrule
Gemini 3 Flash       & 95.2 & 89.6 & 87.1 & 85.4 \\
GPT-5.4               & 90.4 & 81.5 & 77.0 & 83.8 \\
Gemini 3.1 Flash-Lite & 87.4 & 77.7 & 73.5 & 63.3 \\
GPT-5 mini            & 87.0 & 83.1 & 78.5 & 48.9 \\
GPT-5.4 mini          & 85.8 & 79.1 & 74.7 & 77.1 \\
Qwen3-VL-235B$^\ddagger$ & 78.8 & 77.0 & 54.4 & 50.5 \\
\addlinespace
UI-TARS-1.5-7B$^\dagger$ & --- & --- & --- & 12.6 \\
\bottomrule
\end{tabular}
\caption{ComponentBench-Full pass rates (\%) by model and observation/action space. All runs use a 20-step budget. Browser-Use is a tool-rich reference regime with DOM access; AX-tree, SoM, and Pixel share the same harness and differ only in observation and action space. Task-resampling bootstrap 95\% CIs ($B{=}10{,}000$) have half-width $\leq$1.9\% per cell. $^\dagger$Native pixel-only harness with coordinate prediction. $^\ddagger$Browser-Use served via Amazon Bedrock (precision unspecified); other Qwen cells use local vLLM FP8 (Appendix~\ref{app:setup}).}
\label{tab:main-results}
\end{table}

Table~\ref{tab:main-results} presents the main v1 results. We highlight three observations.

\paragraph{Observation/action space can shift pass rates by over 30\% within a single model.} The most dramatic example is GPT-5~mini, which drops from 83.1\% in AX-tree to 48.9\% in Pixel---a 34.2\% swing within the same harness, where only the observation and action space changes (87.0\% to 48.9\% including the tool-rich Browser-Use regime). Every model evaluated in multiple spaces shows a substantial gap between its best and worst space, confirming that the interface through which an agent encounters a component is not merely a convenience choice but a major determinant of success.

\paragraph{The benefit of structured aids is model-dependent.} For models that are weaker at direct screenshot-based interaction, SoM provides a clear advantage over Pixel: GPT-5~mini gains 29.6\% and Gemini~3.1~Flash-Lite gains 10.2\%. But for two of the six models evaluated in both SoM and Pixel---GPT-5.4~mini and GPT-5.4---the ordering \emph{reverses}: Pixel outperforms SoM by 2.4\% and 6.8\% (paired bootstrap 95\% CIs [1.0, 3.9] and [5.3, 8.3]). For GPT-5.4, Pixel (83.8\%) also outperforms AX-tree (81.5\%). Gemini~3~Flash sits near the crossover, with only a 1.7\% SoM advantage. This pattern suggests that structured overlays are not uniformly beneficial: they help weaker models substantially, but for models already strong at direct visual interaction, the added clutter and indirection of SoM can become neutral or slightly harmful.

\paragraph{Both model capability and interface regime have large effects, and the benchmark is far from saturated.} Gemini~3~Flash achieves 85.4\% even in Pixel---the most restrictive observation space---exceeding several weaker models' Browser-Use performance. At the other end, UI-TARS-1.5-7B, a 7B model trained specifically for GUI interaction, achieves 12.6\% in its native pixel-only mode, with 87\% of tasks hitting the 20-step ceiling.

\section{Analysis}

\subsection{Agents solve most tasks but rarely match human efficiency}
\label{sec:step-efficiency}

\begin{table}[t]
\centering
\footnotesize
\setlength{\tabcolsep}{3pt}
\begin{tabular}{@{}lrrrrrrrrrrrr@{}}
\toprule
& \multicolumn{3}{c}{\textbf{Browser-Use}} & \multicolumn{3}{c}{\textbf{AX-tree}} & \multicolumn{3}{c}{\textbf{SoM}} & \multicolumn{3}{c}{\textbf{Pixel}} \\
\cmidrule(lr){2-4}\cmidrule(lr){5-7}\cmidrule(lr){8-10}\cmidrule(lr){11-13}
\textbf{Model} & Pass & $\leq H$ & $\leq$2$H$ & Pass & $\leq H$ & $\leq$2$H$ & Pass & $\leq H$ & $\leq$2$H$ & Pass & $\leq H$ & $\leq$2$H$ \\
\midrule
Gemini 3 Flash    & 95.2 & 56.1 & 85.3 & 89.6 & 74.7 & 82.5 & 87.1 & 73.2 & 81.4 & 85.4 & 65.0 & 77.8 \\
GPT-5.4           & 90.4 & 40.6 & 75.3 & 81.5 & 56.5 & 72.4 & 77.0 & 49.6 & 68.2 & 83.8 & 48.1 & 69.5 \\
Gemini 3.1 FL     & 87.4 & 50.2 & 77.3 & 77.7 & 68.6 & 73.9 & 73.5 & 64.8 & 70.9 & 63.3 & 47.3 & 56.2 \\
GPT-5 mini        & 87.0 & 51.7 & 78.5 & 83.1 & 66.6 & 76.5 & 78.5 & 63.2 & 73.4 & 48.9 & 19.0 & 28.2 \\
GPT-5.4 mini      & 85.8 & 51.0 & 77.9 & 79.1 & 69.0 & 73.8 & 74.7 & 64.0 & 70.6 & 77.1 & 60.2 & 68.9 \\
Qwen3-VL-235B     & 78.8 & 39.7 & 70.1 & 77.0 & 66.1 & 73.2 & 54.4 & 41.1 & 48.4 & 50.5 & 29.3 & 38.1 \\
UI-TARS$^\dagger$ & --- & --- & --- & --- & --- & --- & --- & --- & --- & 12.6 & 9.0 & 10.5 \\
\bottomrule
\end{tabular}
\caption{Step efficiency (\% of 2,910 tasks). $\leq H$/$\leq$2$H$: solved within 1/2$\times$ human steps, where $H=\max(\text{human steps},1)$; hover-only references count as $H{=}1$. Browser-Use ``steps'' are model turns of up to 4 primitive actions (page-load turn excluded), so its efficiency columns are optimistic. $^\dagger$Native pixel-only harness. Full $\leq$3$H$/5$H$ breakdown in Appendix~\ref{app:full-step-efficiency}.}
\label{tab:step-efficiency}
\end{table}

Table~\ref{tab:step-efficiency} reports not only whether tasks are solved, but whether they are solved \emph{efficiently}. The gap between Pass and $\leq H$ is the \emph{efficiency gap}: tasks solvable but requiring more effort than a human. Because a Browser-Use step can bundle up to four primitive actions, its $\leq H$ columns are an optimistic turn-level reference rather than a like-for-like action comparison. AX-tree and SoM are the most step-efficient modes (median ratio 1.0$\times$). Browser-Use has the largest gap---Gemini~3~Flash reaches 95.2\% pass but only 56.1\% within human steps---and the gap is most severe for weak model--hard mode combinations: GPT-5~mini Pixel passes 48.9\% but only 19.0\% within human steps (mean ratio 3.6$\times$, p90 = 9.0$\times$). Wall-clock time reinforces this: even the fastest configuration (GPT-5.4~mini SoM) takes 3.7$\times$ as long as the matched human reference, and the slowest (GPT-5~mini Pixel) takes 21.5$\times$ as long, averaging 71.8 seconds per successful task (full time data in the appendix).

\subsection{Browser-Use is a tool-rich reference regime with family-dependent trade-offs}
\label{sec:repr-dependence}

Browser-Use provides an aggregate pass-rate advantage ($+$7.8\% to $+$18.2\% over the mean of other modes), largest for weaker models, but at the cost of lower step efficiency. Part of this advantage comes from its ability to execute up to 4 actions per model turn (up to 80 actions versus 20 for the other modes). The advantage is not uniform, however: across the six models it is strongly family-dependent, $+$29.0\% on Advanced Editors but $-$24.0\% on Drag/Drop. The Layer~3 diagnostic reports explain why: Browser-Use's DOM-level action surface cannot replicate genuine spatial drag operations on tasks like Kanban boards; on this family, GPT-5.4 falls from 70\% in AX-tree and 63\% in Pixel to 55\% in Browser-Use. The CUA gap also scales with intended difficulty ($+$7.2\% on easy tasks, $+$18.2\% on hard).

\subsection{Spatial manipulation components are trivial for humans but hard for agents}
\label{sec:family-analysis}

\begin{table}[t]
\centering
\small
\setlength{\tabcolsep}{3pt}
\begin{tabular}{@{}lrrrr r@{}}
\toprule
\textbf{Family} & \textbf{BU} & \textbf{AX} & \textbf{SoM} & \textbf{Pix} & \textbf{Avg} \\
\midrule
Drag/Drop \& Workspace       & 29.7 & 72.1 & 31.8 & 57.4 & 47.7 \\
Continuous Precision          & 74.0 & 53.8 & 53.9 & 56.2 & 59.4 \\
Advanced Editors              & 83.6 & 67.8 & 49.7 & 46.4 & 61.9 \\
Date \& Time                  & 83.4 & 76.2 & 65.8 & 59.4 & 71.2 \\
Disclosure \& Progressive     & 80.1 & 76.0 & 64.1 & 68.3 & 72.1 \\
\addlinespace
Discrete Choice               & 97.2 & 85.7 & 83.3 & 79.0 & 86.3 \\
Overlays \& Transient UI      & 96.4 & 91.0 & 88.6 & 80.7 & 89.2 \\
Command \& Navigation         & 97.3 & 96.1 & 89.9 & 83.1 & 91.6 \\
\bottomrule
\end{tabular}
\caption{Pass rate (\%) by family and mode, averaged across all six models (excluding UI-TARS). Top: five hardest families; bottom: three easiest. Full breakdown in Figure~\ref{fig:family-heatmap} (Appendix).}
\label{tab:family-passrate}
\end{table}

Table~\ref{tab:family-passrate} summarizes family-level pass rates. Families span from Command \& Navigation (91.6\% average) to Drag/Drop (47.7\%), and no single observation/action space is universally best: Drag/Drop favors Pixel over Browser-Use, while Advanced Editors strongly favor Browser-Use. The full per-model heatmap is in the appendix (Figure~\ref{fig:family-heatmap}).

The Layer~3 diagnostic reports reveal distinct mechanistic patterns behind these family-level differences: on \emph{editable data grids} agents identify the correct row yet fail to enter the widget's editor mode or commit through its expected control (an \emph{interaction knowledge} failure rather than a grounding failure); on \emph{rich text editors} they find the toolbar but cannot establish and hold transient contenteditable selections; on \emph{context menus} the difficulty is not locating the target but controlling the transient overlay after the right-click. Section~\ref{sec:failure-taxonomy} quantifies these mechanisms across all failed traces.

A striking \emph{difficulty inversion} emerges when comparing human and agent difficulty: nine canonical types require $\leq$2 human steps yet achieve $<$60\% agent pass rate---resizable\_columns (24.4\%), window\_splitter (38.3\%), slider\_range (39.9\%), and six other spatial manipulation tasks (see Figure~\ref{fig:inversion} in the appendix and Table~\ref{tab:hardest-types}). These are trivial mouse gestures for humans but among the hardest interactions for agents. The Layer~3 report for window splitters explains that the difficulty is ``compressing perception, motor control, and verification into one thin affordance'': every task has maximal precision requirement, but outcomes depend on whether the separator is even exposed as a real control in the agent's observation space. No component type shows the reverse pattern (hard for humans, easy for agents).

\subsection{A trace-grounded failure taxonomy}
\label{sec:failure-taxonomy}

To move from family-level pass rates to failure \emph{mechanisms}, we assign every failed trace a primary diagnostic category supported by trace evidence, using two complementary labelings (Table~\ref{tab:failure-taxonomy}): a \emph{deterministic trace-feature taxonomy} over all 8,864 failed Pixel/SoM/AX-tree traces across the five models with complete BrowserGym trace logs (each trace is parsed into its action sequence and mapped from a component-driven prior refined by trace evidence), and the \emph{Layer-2 diagnostic labels} (an LLM reading of all 2,752 failed runs of one model) mapped onto the same categories. Methodological details and adversarially reviewed case studies appear in Appendix~\ref{app:failure-analysis}.

\begin{table}[t]
\centering
\small
\setlength{\tabcolsep}{4pt}
\begin{tabular}{@{}lrr@{}}
\toprule
\textbf{Failure category} & \textbf{Deterministic} & \textbf{Layer-2} \\
 & (5 models, $n$=8{,}864) & (1 model, $n$=2{,}752) \\
\midrule
Continuous calibration error            & 20.2\% & --- \\
Transient state loss                    & 19.9\% & 15.0\% \\
Missing commit or confirmation          & 11.6\% & 20.0\% \\
Target acquisition / wrong instance     & 11.2\% & 31.6\% \\
Repetition / no-progress loop           & 11.2\% & 5.6\% \\
Widget-specific procedure missing       & 9.6\%  & 24.1\% \\
Drag execution failure                  & 4.3\%  & --- \\
Semantic value error                    & 3.6\%  & 3.3\% \\
Other / unclear                         & 8.4\%  & 0.3\% \\
\bottomrule
\end{tabular}
\caption{Failure taxonomy over all failed BrowserGym-mode traces, under two complementary labelings: a deterministic trace-feature pass over five models, and Layer-2 LLM diagnostic labels for Gemini~3.1~Flash-Lite mapped onto the same categories. The two labelings surface overlapping high-frequency mechanism families, though their percentages are not directly comparable (different model sets and schemas). Continuous-calibration and drag-execution failures are not represented as separate categories in the original Layer-2 schema, hence absent from that column.}
\label{tab:failure-taxonomy}
\end{table}

The two labelings surface overlapping high-frequency mechanism families, though their percentages are not directly comparable: \emph{continuous calibration errors} (the agent engages the right slider or meter but cannot map pointer movement to the required value, overshooting and undershooting until timeout), \emph{transient state loss} (an opened popover, editor mode, or selection is lost before commit), \emph{target acquisition / wrong-instance errors}, \emph{missing commit actions}, and \emph{missing widget-specific procedures}. Two cross-cutting observations sharpen the earlier findings. First, slider and meter failures are usually \emph{not} instruction-understanding failures: agents locate the correct control but cannot calibrate the continuous value, which explains why these tasks are trivial for humans (one drag) yet resistant to added reasoning. Second, a no-progress \emph{loop} is a symptom rather than a root cause: 55.8\% of failed traces end in a repeated-action loop, but it is distributed across every mechanism ($\approx$48--67\% within each category)---agents loop \emph{because} they are stuck on the underlying mechanism.

\subsection{Clutter and spacing disproportionately burden visual agents}
\label{sec:scene-factors}

Scene factors create differential burdens across observation/action spaces (Figure~\ref{fig:scene-factors}, Appendix). Averaged across models, medium clutter drops Pixel by 14.3\% but AX-tree by only 0.2\%; compact spacing drops Pixel by 13.6\% but AX-tree by 4.5\%. The failure taxonomy shows this is mechanistic: clutter manifests as target-acquisition and wrong-instance failures, stressing visual grounding rather than task semantics---which is why Browser-Use, acting on DOM elements, is essentially clutter-immune (medium clutter costs it 0.8\%). The interaction between model capability and scene factors is especially clear for GPT-5.4: at clutter=none, GPT-5.4 Pixel (86.1\%) \emph{outperforms} AX-tree (81.6\%) by 4.5\%, but at clutter=medium the gap reverses and AX-tree (82.3\%) leads Pixel (72.6\%) by 9.7\%---the preferred observation space flips on a single scene factor.

The intended difficulty axes also validate empirically: precision requirement is the most predictive axis ($r = +0.44$ with failure rate), and the prediction is mode-dependent ($r = +0.41$ for Pixel vs.\ $r = +0.05$ for Browser-Use on depth/layering). Difficulty tiers decline monotonically: L0 = 87.3\%, L3 = 65.2\%, with the AX-tree--Pixel gap widening from 4.6\% (L0) to 21.9\% (L3). Across all models, a task-level asymmetry also emerges: 123 tasks pass reliably in AX-tree (mean pass rate $>$0.7) while failing in Pixel (mean $<$0.3), whereas only 52 show the reverse---the asymmetry between structured and visual observation is large and directional.

\subsection{ComponentBench-Core as a pressure test}
\label{sec:core}

ComponentBench-Core is derived from the full suite using the evidence described in Section~\ref{sec:full-to-core}: realized-difficulty audits, Layer~2/3 diagnostic reports, and scene-factor interactions merge overlapping canonical types into 19 interaction-centered generation units, each contributing 48 regenerated hard-only tasks (912 total). Core is not a subset of Full: its tasks are newly generated, dropping saturated easy families and concentrating benchmark mass on the interaction patterns that still separate current agents.

\begin{table}[h]
\centering
\small
\setlength{\tabcolsep}{4pt}
\begin{tabular}{@{}llrrrr@{}}
\toprule
\textbf{Model} & \textbf{Mode} & \textbf{Pass} & \boldmath$\leq H$ & \boldmath$\leq 2H$ & \boldmath$\leq 3H$ \\
\midrule
Gemini 3 Flash & Browser-Use & 84.5 & 51.5 & 71.5 & 78.5 \\
Gemini 3 Flash & Pixel       & 60.9 & 30.5 & 51.0 & 56.0 \\
GPT-5.4 mini   & Browser-Use & 57.8 & 36.8 & 51.2 & 55.3 \\
GPT-5.4 mini   & Pixel       & 37.7 & 22.0 & 32.1 & 34.1 \\
\addlinespace
Opus 4.6       & Pixel       & 65.4 & 34.1 & 53.8 & 59.4 \\
\bottomrule
\end{tabular}
\caption{ComponentBench-Core results (\% of 912 tasks). Same metrics (and Browser-Use turn-level caveat) as Table~\ref{tab:step-efficiency}, on the hard-only \textsc{Core} suite. Pass rates drop 10--39\% from \textsc{Full}, confirming that \textsc{Core} concentrates diagnostic mass on unresolved interaction families. Opus~4.6 is evaluated only on \textsc{Core} Pixel.}
\label{tab:core-results}
\end{table}

Table~\ref{tab:core-results} presents the results on \textsc{Core}. Pass rates drop substantially relative to \textsc{Full}: Gemini~3~Flash Browser-Use falls from 95.2\% to 84.5\% ($-$10.7\%), Gemini~3~Flash Pixel from 85.4\% to 60.9\% ($-$24.5\%), GPT-5.4~mini Browser-Use from 85.8\% to 57.8\% ($-$28.0\%), and GPT-5.4~mini Pixel from 77.1\% to 37.7\% ($-$39.4\%). The drop is largest for the weakest combination: \textsc{Core} disproportionately stresses the modes and models already borderline on \textsc{Full}.

To test whether \textsc{Core} remains diagnostic for frontier models not evaluated on \textsc{Full}, we additionally run Opus~4.6 \citep{anthropic2026opus46} on \textsc{Core} Pixel. It achieves 65.4\%---the highest pixel-only result on the hard suite, surpassing Gemini~3~Flash Pixel (60.9\%)---but still leaves over a third of tasks unsolved, with only 34.1\% solved within the human step budget and a 9.5$\times$ time-to-human ratio (67.4 vs.\ 7.1 mean seconds per successful task). \textsc{Core} is therefore not an artifact of weaker models: it stays challenging even for a frontier computer-use model. Human traces for \textsc{Core} average 5.2 normalized steps versus 2.7 for \textsc{Full}, reflecting the \textsc{Core} tasks' structural complexity. Where \textsc{Full} suits broad exploration, \textsc{Core} is a compact stress test for the hardest families.

\section{Limitations and scope}

ComponentBench targets recurring component families from WAI-ARIA patterns and three major production libraries, not every bespoke widget in the wild; CAPTCHA-like tasks are excluded. The benchmark is web-first (Next.js on Chromium), though the ontology is designed to port to desktop and mobile. It measures component-level competence, not long-horizon planning; validating how component-level scores predict end-to-end workflow success is future work.

The realized-difficulty audit is a hybrid system: some quantities are measured from the rendered DOM, others approximated from type-level metadata. The primary human reference traces come from two passes by a single annotator---a practical reference, not a proof of near-optimality; a validation study with two additional annotators (Appendix~\ref{app:human-validation}) shows inter-annotator variation (1.05$\times$ per-task) is small relative to the agent--human gap (1.27--3.02$\times$), though agents with tree- or DOM-level tools may admit shorter non-visual paths. Both should be treated as structured operationalizations, not oracles.

Our experiments cover seven models on the full suite and one additional frontier model (Opus~4.6) on Core---a broad but not exhaustive snapshot. The main tables report a single run per model--mode combination; in a repeated-run study on a 278-task subset (Appendix~\ref{app:stability}) the observed run-to-run deviation was at most 1.4\% with no ordering changes, though task-level outcomes churn on borderline tasks.

Because task specifications were generated with GPT-5.2~Pro and pages implemented with Claude~Code, the benchmark may contain generator-specific regularities. Three design choices mitigate this risk: every task is scored by a deterministic programmatic verifier rather than LLM judgment, the generation prompt enforces combinatorial diversity over a fixed ontology (Appendix~\ref{app:task-diversity}), and the human recording passes (including the annotators of Appendix~\ref{app:human-validation}) served as quality checks that surfaced no broken, ambiguous, or miswired tasks. We do not, however, audit exhaustively for contamination.

Additionally, GPT models were accessed via the chat completions API rather than OpenAI's Operator or Computer Use API, Opus~4.6 through a custom harness rather than Anthropic's full computer-use environment, and BrowserGym renders no visible cursor---so our results may underestimate native-interface performance.

\section{Conclusion}

We presented ComponentBench, a component-level benchmark for diagnosing where computer-use agents fail on modern web UIs. Across 97 canonical component types, 2,910 tasks, and four observation regimes, representation, efficiency, and visual context all critically shape agent performance, and a trace-grounded failure taxonomy ties these effects to concrete mechanisms---helping localize the component-level causes of workflow failures.

\section*{Acknowledgments}
Model API access for the experiments in this paper was provided by the Amazon AGI SF Lab. Experiments were run on the Duke Computer Science cluster. We thank Jiacheng Sang and Xunjian Yin for contributing human reference annotations and for helpful advice.

\bibliographystyle{colm2026_conference}
\bibliography{componentbench_refs}

\begin{thebibliography}{36}
\providecommand{\natexlab}[1]{#1}
\providecommand{\url}[1]{\texttt{#1}}
\expandafter\ifx\csname urlstyle\endcsname\relax
  \providecommand{\doi}[1]{doi: #1}\else
  \providecommand{\doi}{doi: \begingroup \urlstyle{rm}\Url}\fi

\bibitem[Abhyankar et~al.(2025)Abhyankar, Qi, and
  Zhang]{abhyankar2025osworldhuman}
Reyna Abhyankar, Qi~Qi, and Yiying Zhang.
\newblock {OSWorld-Human}: Benchmarking the efficiency of computer-use agents,
  2025.
\newblock URL \url{https://arxiv.org/abs/2506.16042}.

\bibitem[{Ant Group}(2026)]{antdesign_components}
{Ant Group}.
\newblock Ant design components overview.
\newblock \url{https://ant.design/components/overview/}, 2026.
\newblock Accessed 2026-03-25.

\bibitem[{Anthropic}(2026{\natexlab{a}})]{anthropic2026computeruse}
{Anthropic}.
\newblock Computer use tool --- {Claude} {API} documentation.
\newblock
  \url{https://docs.anthropic.com/en/docs/agents-and-tools/computer-use},
  2026{\natexlab{a}}.
\newblock Accessed 2026-03-25.

\bibitem[{Anthropic}(2026{\natexlab{b}})]{anthropic2026opus46}
{Anthropic}.
\newblock Introducing {Claude Opus} 4.6.
\newblock \url{https://www.anthropic.com/news/claude-opus-4-6},
  2026{\natexlab{b}}.
\newblock Accessed 2026-07-24.

\bibitem[{Anthropic}(2026{\natexlab{c}})]{anthropic2026opus48}
{Anthropic}.
\newblock Introducing {Claude Opus} 4.8.
\newblock \url{https://www.anthropic.com/news/claude-opus-4-8},
  2026{\natexlab{c}}.
\newblock Accessed 2026-07-24.

\bibitem[Bai et~al.(2025)]{wang2025qwen3vl}
Shuai Bai et~al.
\newblock Qwen3-{VL} technical report, 2025.
\newblock URL \url{https://arxiv.org/abs/2511.21631}.

\bibitem[{Chakra UI}(2026)]{chakraui_components}
{Chakra UI}.
\newblock Chakra {UI} components.
\newblock \url{https://www.chakra-ui.com/docs/components/concepts/overview},
  2026.
\newblock Accessed 2026-07-11.

\bibitem[Cheng et~al.(2024)Cheng, Sun, Chu, Xu, Li, Zhang, and
  Wu]{cheng2024seeclick}
Kanzhi Cheng, Qiushi Sun, Yougang Chu, Fangzhi Xu, Yantao Li, Jianbing Zhang,
  and Zhiyong Wu.
\newblock Seeclick: Harnessing gui grounding for advanced visual gui agents.
\newblock In \emph{Proceedings of the 62nd Annual Meeting of the Association
  for Computational Linguistics}, 2024.
\newblock URL \url{https://arxiv.org/abs/2401.10935}.

\bibitem[de~Chezelles et~al.(2024)de~Chezelles, Gasse, Drouin, Caccia,
  Boisvert, Thakkar, Marty, Assouel, Shayegan, Jang, L{\`u}, Yoran, Kong, Xu,
  Reddy, Cappart, Neubig, Salakhutdinov, Chapados, and
  Lacoste]{dechezelles2024browsergym}
Thibault Le~Sellier de~Chezelles, Maxime Gasse, Alexandre Drouin, Massimo
  Caccia, L{\'e}o Boisvert, Megh Thakkar, Tom Marty, Rim Assouel, Sahar~Omidi
  Shayegan, Lawrence~Keunho Jang, Xing~Han L{\`u}, Ori Yoran, Dehan Kong,
  Frank~F. Xu, Siva Reddy, Quentin Cappart, Graham Neubig, Ruslan
  Salakhutdinov, Nicolas Chapados, and Alexandre Lacoste.
\newblock The browsergym ecosystem for web agent research, 2024.
\newblock URL \url{https://arxiv.org/abs/2412.05467}.

\bibitem[Deng et~al.(2023)Deng, Gu, Zheng, Chen, Stevens, Wang, Sun, and
  Su]{deng2023mind2web}
Xiang Deng, Yu~Gu, Boyuan Zheng, Shijie Chen, Samuel Stevens, Boshi Wang, Huan
  Sun, and Yu~Su.
\newblock Mind2web: Towards a generalist agent for the web, 2023.
\newblock URL \url{https://arxiv.org/abs/2306.06070}.

\bibitem[Fitts(1954)]{fitts1954}
Paul~M. Fitts.
\newblock The information capacity of the human motor system in controlling the
  amplitude of movement.
\newblock \emph{Journal of Experimental Psychology}, 47\penalty0 (6):\penalty0
  381--391, 1954.
\newblock \doi{10.1037/h0055392}.

\bibitem[{Google}(2026)]{google2026gemini31flashlite}
{Google}.
\newblock Gemini 3.1 {Flash-Lite}: Built for intelligence at scale.
\newblock
  \url{https://blog.google/innovation-and-ai/models-and-research/gemini-models/gemini-3-1-flash-lite/},
  2026.
\newblock Accessed 2026-07-24.

\bibitem[{Google DeepMind}(2025)]{google2025gemini3flash}
{Google DeepMind}.
\newblock Gemini 3 flash model card.
\newblock
  \url{https://storage.googleapis.com/deepmind-media/Model-Cards/Gemini-3-Flash-Model-Card.pdf},
  2025.
\newblock Accessed 2026-03-30.

\bibitem[He et~al.(2024)He, Yao, Ma, Yu, Dai, Zhang, Lan, and
  Yu]{he2024webvoyager}
Hongliang He, Wenlin Yao, Kaixin Ma, Wenhao Yu, Yong Dai, Hongming Zhang,
  Zhenzhong Lan, and Dong Yu.
\newblock Webvoyager: Building an end-to-end web agent with large multimodal
  models, 2024.
\newblock URL \url{https://arxiv.org/abs/2401.13919}.

\bibitem[Koh et~al.(2024)Koh, Lo, Jang, Duvvur, Lim, Huang, Neubig, Zhou,
  Salakhutdinov, and Fried]{koh2024visualwebarena}
Jing~Yu Koh, Robert Lo, Lawrence Jang, Vikram Duvvur, Ming~Chong Lim, Po-Yu
  Huang, Graham Neubig, Shuyan Zhou, Ruslan Salakhutdinov, and Daniel Fried.
\newblock Visualwebarena: Evaluating multimodal agents on realistic visual web
  tasks, 2024.
\newblock URL \url{https://arxiv.org/abs/2401.13649}.

\bibitem[Li \& Waldo(2024)Li and Waldo]{li2024websuite}
Eric Li and Jim Waldo.
\newblock Websuite: Systematically evaluating why web agents fail, 2024.
\newblock URL \url{https://arxiv.org/abs/2406.01623}.

\bibitem[Li et~al.(2025)Li, Meng, Lin, Luo, Tian, Ma, Huang, and
  Chua]{li2025screenspotpro}
Kaixin Li, Ziyang Meng, Hongzhan Lin, Ziyang Luo, Yuchen Tian, Jing Ma, Zhiyong
  Huang, and Tat-Seng Chua.
\newblock {ScreenSpot-Pro}: Gui grounding for professional high-resolution
  computer use, 2025.
\newblock URL \url{https://arxiv.org/abs/2504.07981}.

\bibitem[Liu et~al.(2018)Liu, Guu, Pasupat, Shi, and Liang]{liu2018workflow}
Evan~Zheran Liu, Kelvin Guu, Panupong Pasupat, Tianlin Shi, and Percy Liang.
\newblock Reinforcement learning on web interfaces using workflow-guided
  exploration.
\newblock In \emph{International Conference on Learning Representations}, 2018.
\newblock URL \url{https://arxiv.org/abs/1802.08802}.

\bibitem[{Mantine}(2026)]{mantine_components}
{Mantine}.
\newblock Mantine core components.
\newblock \url{https://mantine.dev/core/package/}, 2026.
\newblock Accessed 2026-03-25.

\bibitem[{Microsoft}(2026{\natexlab{a}})]{fluentui_components}
{Microsoft}.
\newblock Fluent {UI} {React} components.
\newblock \url{https://react.fluentui.dev/}, 2026{\natexlab{a}}.
\newblock Accessed 2026-03-25.

\bibitem[{Microsoft}(2026{\natexlab{b}})]{playwright_eval}
{Microsoft}.
\newblock Evaluating javascript --- playwright documentation.
\newblock \url{https://playwright.dev/docs/evaluating}, 2026{\natexlab{b}}.
\newblock Accessed 2026-03-25.

\bibitem[{MUI}(2026)]{mui_components}
{MUI}.
\newblock Mui components documentation.
\newblock \url{https://mui.com/components/}, 2026.
\newblock Accessed 2026-03-25.

\bibitem[M{\"u}ller \& Zu{\ss}(2024)M{\"u}ller and Zu{\ss}]{browseruse2024}
Magnus M{\"u}ller and Gregor Zu{\ss}.
\newblock Browser-use: Make websites accessible for {AI} agents.
\newblock \url{https://github.com/browser-use/browser-use}, 2024.
\newblock MIT License, v0.12.

\bibitem[{OpenAI}(2025{\natexlab{a}})]{openai2025gpt5}
{OpenAI}.
\newblock {OpenAI GPT-5} system card, 2025{\natexlab{a}}.
\newblock URL \url{https://arxiv.org/abs/2601.03267}.

\bibitem[{OpenAI}(2025{\natexlab{b}})]{openai2025operator}
{OpenAI}.
\newblock Introducing operator.
\newblock \url{https://openai.com/index/introducing-operator/},
  2025{\natexlab{b}}.
\newblock Accessed 2026-03-25.

\bibitem[{OpenAI}(2026{\natexlab{a}})]{openai2026computeruse}
{OpenAI}.
\newblock Computer use --- {OpenAI} {API} documentation.
\newblock \url{https://platform.openai.com/docs/guides/tools-computer-use},
  2026{\natexlab{a}}.
\newblock Accessed 2026-07-11.

\bibitem[{OpenAI}(2026{\natexlab{b}})]{openai2026gpt54}
{OpenAI}.
\newblock Introducing {GPT-5.4}.
\newblock \url{https://openai.com/index/introducing-gpt-5-4/},
  2026{\natexlab{b}}.
\newblock Accessed 2026-03-30.

\bibitem[{OpenAI}(2026{\natexlab{c}})]{openai2026gpt54mini}
{OpenAI}.
\newblock Introducing {GPT-5.4} mini and nano.
\newblock \url{https://openai.com/index/introducing-gpt-5-4-mini-and-nano/},
  2026{\natexlab{c}}.
\newblock Accessed 2026-07-24.

\bibitem[Qin et~al.(2025)Qin, Ye, Fang, Wang, Liang, Tian, Zhang, Li, Li,
  Huang, Zhong, Li, et~al.]{qin2025uitars}
Yujia Qin, Yining Ye, Junjie Fang, Haoming Wang, Shihao Liang, Shizuo Tian,
  Junda Zhang, Jiahao Li, Yunxin Li, Shijue Huang, Wanjun Zhong, Kuanye Li,
  et~al.
\newblock {UI-TARS}: Pioneering automated {GUI} interaction with native agents,
  2025.
\newblock URL \url{https://arxiv.org/abs/2501.12326}.

\bibitem[Srivastava et~al.(2025)Srivastava, Li, Chang, Garg, Kaur, Lee, Li,
  Mao, Cases, Xie, and Qi]{srivastava2025warc}
Sanjari Srivastava, Gang Li, Cheng Chang, Rishu Garg, Manpreet Kaur,
  Charlene~Y. Lee, Yuezhang Li, Yining Mao, Ignacio Cases, Yanan Xie, and Peng
  Qi.
\newblock Warc-bench: Web archive based benchmark for gui subtask executions,
  2025.
\newblock URL \url{https://arxiv.org/abs/2510.09872}.

\bibitem[{Tailwind Labs}(2026)]{headlessui_components}
{Tailwind Labs}.
\newblock Headless {UI} components.
\newblock \url{https://headlessui.com/}, 2026.
\newblock Accessed 2026-03-25.

\bibitem[Ullrich et~al.(2025)Ullrich, Su, Shi, Subramonian, Bar, Evtimov,
  Tsilivis, Balestriero, Kempe, and Ibrahim]{ullrich2025openapps}
Karen Ullrich, Jingtong Su, Claudia Shi, Arjun Subramonian, Amir Bar, Ivan
  Evtimov, Nikolaos Tsilivis, Randall Balestriero, Julia Kempe, and Mark
  Ibrahim.
\newblock {OpenApps}: Simulating environment variations to measure {UI}-agent
  reliability, 2025.
\newblock URL \url{https://arxiv.org/abs/2511.20766}.

\bibitem[{World Wide Web Consortium}(2026)]{w3c_apg}
{World Wide Web Consortium}.
\newblock Aria authoring practices guide (apg).
\newblock \url{https://www.w3.org/WAI/ARIA/apg/}, 2026.
\newblock Accessed 2026-03-25.

\bibitem[Xie et~al.(2024)Xie, Zhang, Chen, Li, Zhao, Cao, Hua, Cheng, Shin,
  Lei, Liu, Xu, Zhou, Savarese, Xiong, Zhong, and Yu]{xie2024osworld}
Tianbao Xie, Danyang Zhang, Jixuan Chen, Xiaochuan Li, Siheng Zhao, Ruisheng
  Cao, Toh~Jing Hua, Zhoujun Cheng, Dongchan Shin, Fangyu Lei, Yitao Liu,
  Yiheng Xu, Shuyan Zhou, Silvio Savarese, Caiming Xiong, Victor Zhong, and Tao
  Yu.
\newblock Osworld: Benchmarking multimodal agents for open-ended tasks in real
  computer environments, 2024.
\newblock URL \url{https://arxiv.org/abs/2404.07972}.

\bibitem[Xue et~al.(2025)Xue, Qi, Shi, Song, Gou, Song, Sun, and
  Su]{xue2025illusion}
Tianci Xue, Weijian Qi, Tianneng Shi, Chan~Hee Song, Boyu Gou, Dawn Song, Huan
  Sun, and Yu~Su.
\newblock An illusion of progress? assessing the current state of web agents,
  2025.
\newblock URL \url{https://arxiv.org/abs/2504.01382}.

\bibitem[Zhou et~al.(2023)Zhou, Xu, Zhu, Zhou, Lo, Sridhar, Cheng, Ou, Bisk,
  Fried, Alon, and Neubig]{zhou2023webarena}
Shuyan Zhou, Frank~F. Xu, Hao Zhu, Xuhui Zhou, Robert Lo, Abishek Sridhar,
  Xianyi Cheng, Tianyue Ou, Yonatan Bisk, Daniel Fried, Uri Alon, and Graham
  Neubig.
\newblock Webarena: A realistic web environment for building autonomous agents,
  2023.
\newblock URL \url{https://arxiv.org/abs/2307.13854}.

\end{thebibliography}

\appendix
\preto{\section}{\FloatBarrier}

\section{Benchmark construction details}
\label{app:construction}

\subsection{Task specification schema}

Each task is specified in YAML with the following fields: \texttt{id}, \texttt{canonical\_type}, \texttt{implementation\_source} (antd/mui/mantine/external), \texttt{task\_template}, \texttt{browsergym\_goal} (natural-language instruction), \texttt{scene\_context} (8 controlled factors), \texttt{difficulty} (bucket, tier, and 7 axis ratings with justification), \texttt{success\_trigger} (human-readable criteria and canonical predicate), \texttt{negative\_cases}, and \texttt{expected\_interaction\_path}. Scene factors and their realized levels are: \texttt{theme} (light/dark), \texttt{spacing} (comfortable/compact), \texttt{layout} (8 levels; isolated\_card, form\_section, dashboard, settings\_panel, \ldots), \texttt{placement} (center plus four off-center corners), \texttt{scale} (default/small/large), \texttt{instances} (1--10), \texttt{guidance} (text/visual/mixed), and \texttt{clutter} (none/low/medium/high). Table~\ref{tab:task-diversity} reports the realized distribution over these levels.

\subsection{LLM-assisted construction pipeline}

GPT-5.2~Pro generated 30 task specifications per canonical type, including templates, intended difficulty labels, scene factors, success triggers, and negative cases. Claude~Code then implemented those specifications as real interactive Next.js pages. Each task was subsequently executed by a human and recorded as a reference trajectory. This recording stage also served as the final quality-control pass: tasks that were broken, miswired, or did not match their intended specification after implementation were corrected before inclusion.

\subsection{Task generation prompt}
\label{app:generation-prompt}

For each canonical component type, GPT-5.2~Pro received a structured prompt together with a research CSV file mapping the component to its available library implementations, supported interaction patterns, and difficulty considerations. The prompt specified the following constraints:

\paragraph{Single-component focus and realistic intents.} Each task must target exactly one primary component whose state determines success. Instructions must read as plausible micro-user intents (e.g., ``Set the Price range slider to \$20--\$80''), not benchmark jargon. Internal tolerances, checker rules, and implementation details are excluded from the agent-facing instruction.

\paragraph{Structured output schema.} Each task specification includes: a stable ID, the canonical type, implementation source and variant, a primary task template (from the 24 defined templates), the complete scene context (all 8 factors), difficulty ratings (bucket, tier, and all 7 axis ratings with justification), a detailed setup description of the rendered page, an explicit success trigger with both human-readable conditions and a machine-friendly canonical predicate (including predicate type, target state, tolerance, confirmation requirements, and correct-instance requirements), a list of negative cases, and an expected interaction path for debugging.

\paragraph{Coverage and balance constraints.} The prompt requires exactly 30 tasks per component type, with a fixed difficulty distribution of 10 easy, 10 medium, and 10 hard tasks. When a component is supported by all three primary libraries, tasks are split 10/10/10 across Ant Design, MUI, and Mantine with balanced per-library difficulty. The prompt also enforces minimum coverage of scene-factor variations: at least 3 tasks with dark theme, at least 3 with compact spacing, at least 4 with clutter, at least 4 with multiple instances (if meaningful for the component), and at least 7 distinct task templates per component type.

\paragraph{Difficulty calibration.} The prompt defines a ``default-first'' principle: easy tasks use library defaults and simple contexts, while hardness comes from realistic variations---scene factors (compact spacing, clutter, multiple instances), component feature variants (toggling sub-controls, enabling search, restricting input), and within-component depth (nested overlays, multi-step navigation). Adversarial tricks such as invisible elements or overlapping click traps are explicitly prohibited.

\subsection{Programmatic verification details}

At the environment level, task termination is deterministic: the page's programmatic verifier checks whether the committed end state satisfies the task-specific predicate, and success is signaled by presenting a DOM element (\texttt{\#cb-success-banner}). For tasks requiring explicit commit actions (Apply, Save, Confirm), the verifier checks only the post-commit state. The YAML enumerates negative cases so that nearby but incorrect states do not count as success.

\subsection{Task diversity summary}
\label{app:task-diversity}

Because the tasks are authored with LLM assistance, a natural concern is that they could collapse onto a few repeated patterns. The generation pipeline explicitly enforces combinatorial coverage (Appendix~\ref{app:generation-prompt}); Table~\ref{tab:task-diversity} summarizes the realized distribution of the 2,910 Full tasks over libraries, templates, difficulty, and scene factors. All 97 canonical types contribute exactly 30 tasks; all 24 templates and all levels of every scene factor are exercised, with deliberately skewed marginals (e.g., most tasks use the default clean context, while $\sim$40\% carry at least one added burden such as clutter, compact spacing, dark theme, or multiple instances).

\begin{table}[h]
\centering
\small
\setlength{\tabcolsep}{4pt}
\begin{tabular}{@{}llr@{}}
\toprule
\textbf{Dimension} & \textbf{Levels (count)} & \textbf{Coverage} \\
\midrule
Canonical type & 97 types $\times$ 30 tasks & 100\% \\
Family & 14 families (120--300 tasks each) & 100\% \\
Library & antd 1{,}000 / mui 910 / mantine 970 / external 30 & 4 \\
Task template & 24 templates; most-used \texttt{match\_reference} (351), & 24/24 \\
 & least-used \texttt{file\_manage} (10) & \\
Difficulty bucket & easy 972 / mid 985 / hard 953 & 3/3 \\
Difficulty tier & L0 789 / L1 1{,}050 / L2 818 / L3 253 & 4/4 \\
Theme & light 2{,}600 / dark 310 & 2/2 \\
Spacing & comfortable 2{,}595 / compact 315 & 2/2 \\
Scale & default 2{,}634 / small 272 / large 4 & 3/3 \\
Clutter & none 1{,}751 / low 666 / medium 339 / high 154 & 4/4 \\
Instances & 1: 2{,}266 / 2: 398 / 3: 206 / $\geq$4: 40 & full \\
Layout & 8 layouts; isolated\_card 1{,}866, form\_section 244, & 8/8 \\
 & dashboard 210, settings\_panel 207, others 383 & \\
Placement & center 2{,}456 / off-center (4 corners) 454 & 5/5 \\
Guidance & text 2{,}476 / visual 250 / mixed 184 & 3/3 \\
\bottomrule
\end{tabular}
\caption{Realized diversity of the 2,910 ComponentBench-Full tasks over libraries, templates, difficulty, and the eight controlled scene factors. The 30 \emph{external} tasks are the markdown-editor tasks, which use the third-party \texttt{@uiw/react-md-editor} because none of the three primary libraries ships a core markdown editor. One additional ad-hoc template variant (\texttt{replace\_code}) is used by a single OTP-input task. The realized difficulty buckets deviate slightly from the prompted 970/970/970 split because two later-added types (breadcrumb, pagination) were generated by a run that did not enforce the per-type balance quota.}
\label{tab:task-diversity}
\end{table}

\section{Realized difficulty details}
\label{app:realized}

\subsection{Replay-based measurement}

The realized audit replays a cleaned human trajectory in Playwright \citep{playwright_eval}, using \texttt{page.evaluate()} to run measurement logic inside the browser page. Features are extracted before and after each action, capturing both the initial state and states revealed through interaction (popovers, nested panels, internal scroll regions). The audit covers all 2,910 Full tasks and all 912 Core tasks.

\subsection{Feature list}

The 24 canonical features are:

\begin{enumerate}
\setlength{\itemsep}{0pt}
\item Minimum target size (px)
\item Target spacing (px)
\item Fitts-style acquisition difficulty \citep{fitts1954}
\item Interactable element density
\item Option/choice count
\item Overlay depth
\item Reference step burden
\item Scroll requirement (boolean + depth)
\item Feedback persistence
\item Reset cost
\item Visible state fraction
\item ARIA richness score
\item Contrast ratio
\item Clutter level
\item Placement offset
\item Instance count
\item Control continuity (metadata-derived)
\item State dimensionality (metadata-derived)
\item Approximate granularity (metadata-derived)
\item Tolerance (metadata-derived)
\item Feedback modality (metadata-derived)
\item Confirmation requirement (metadata-derived)
\item Scroll region depth (metadata-derived)
\item Precision surface type (metadata-derived)
\end{enumerate}

Features 1--16 are measured from the rendered DOM and layout; features 17--24 are derived from type-level metadata.

\subsection{Axis aggregation}

The seven realized axes are computed by weighted averaging of normalized features:
\begin{equation}
s_a(t) = \frac{\sum_{f \in \mathcal{F}_a} w_f \,\hat{f}(t)}{\sum_{f \in \mathcal{F}_a} w_f},
\end{equation}
where $\hat{f}(t)$ is the min-max normalized feature value and the current implementation uses unit weights ($w_f = 1$). Continuous scores are discretized with frozen, versioned cutpoints into 1--5 ratings.

\section{Full step efficiency table}
\label{app:full-step-efficiency}

\begin{table}[h]
\centering
\small
\setlength{\tabcolsep}{4pt}
\begin{tabular}{@{}llrrrrr@{}}
\toprule
\textbf{Model} & \textbf{Mode} & \textbf{Pass} & \boldmath$\leq H$ & \boldmath$\leq 2H$ & \boldmath$\leq 3H$ & \boldmath$\leq 5H$ \\
\midrule
Gemini 3 Flash       & Browser-Use & \textbf{95.2} & 56.1 & 85.3 & 89.1 & 91.9 \\
                      & AX-tree     & 89.6 & \textbf{74.7} & 82.5 & 84.7 & 87.4 \\
                      & SoM         & 87.1 & 73.2 & 81.4 & 83.6 & 85.2 \\
                      & Pixel       & 85.4 & 65.0 & 77.8 & 80.9 & 83.6 \\
\midrule
GPT-5.4               & Browser-Use & \textbf{90.4} & 40.6 & 75.3 & 82.9 & 86.8 \\
                      & AX-tree     & 81.5 & \textbf{56.5} & 72.4 & 76.3 & 79.4 \\
                      & SoM         & 77.0 & 49.6 & 68.2 & 72.9 & 75.5 \\
                      & Pixel       & 83.8 & 48.1 & 69.5 & 75.7 & 80.0 \\
\midrule
Gemini 3.1 Flash-Lite & Browser-Use & \textbf{87.4} & 50.2 & 77.3 & 81.1 & 83.8 \\
                      & AX-tree     & 77.7 & \textbf{68.6} & 73.9 & 75.3 & 76.5 \\
                      & SoM         & 73.5 & 64.8 & 70.9 & 72.2 & 72.8 \\
                      & Pixel       & 63.3 & 47.3 & 56.2 & 59.0 & 61.0 \\
\midrule
GPT-5 mini            & Browser-Use & \textbf{87.0} & 51.7 & 78.5 & 82.1 & 84.8 \\
                      & AX-tree     & 83.1 & \textbf{66.6} & 76.5 & 78.8 & 80.6 \\
                      & SoM         & 78.5 & 63.2 & 73.4 & 75.9 & 77.3 \\
                      & Pixel       & 48.9 & 19.0 & 28.2 & 32.7 & 38.7 \\
\midrule
GPT-5.4 mini          & Browser-Use & \textbf{85.8} & 51.0 & 77.9 & 81.6 & 83.7 \\
                      & AX-tree     & 79.1 & \textbf{69.0} & 73.8 & 75.4 & 76.9 \\
                      & SoM         & 74.7 & 64.0 & 70.6 & 72.2 & 73.5 \\
                      & Pixel       & 77.1 & 60.2 & 68.9 & 72.1 & 74.9 \\
\midrule
Qwen3-VL-235B         & Browser-Use & \textbf{78.8} & 39.7 & 70.1 & 74.2 & 77.0 \\
                      & AX-tree     & 77.0 & \textbf{66.1} & 73.2 & 74.8 & 75.7 \\
                      & SoM         & 54.4 & 41.1 & 48.4 & 51.3 & 53.0 \\
                      & Pixel       & 50.5 & 29.3 & 38.1 & 42.3 & 46.3 \\
\midrule
UI-TARS-1.5-7B$^\dagger$ & Pixel & 12.6 & 9.0 & 10.5 & 11.2 & 11.9 \\
\bottomrule
\end{tabular}
\caption{Full step efficiency on ComponentBench-Full (\% of 2,910 tasks). \textbf{Pass}: solved within 20 steps. $\leq H$/$\leq 2H$/$\leq 3H$/$\leq 5H$: solved within 1/2/3/5$\times$ the human step count. \textbf{Bold}: best per model. Browser-Use steps are model turns of up to 4 primitive actions (page-load turn excluded), so its efficiency columns are optimistic. $^\dagger$Native pixel-only interface.}
\label{tab:step-efficiency-full}
\end{table}

\section{Full time efficiency table}
\label{app:time}

\begin{table}[h]
\centering
\small
\setlength{\tabcolsep}{4pt}
\begin{tabular}{@{}llrrrr@{}}
\toprule
\textbf{Model} & \textbf{Mode} & \textbf{Human} & \textbf{Agent} & \textbf{Ratio} & \textbf{Med.\ Agent} \\
\midrule
Gemini 3 Flash       & AX-tree     & 4.5s & 21.5s & 4.7$\times$ & 13.5s \\
                      & SoM         & 4.6s & 22.0s & 4.8$\times$ & 14.4s \\
                      & Pixel       & 4.3s & 27.6s & 6.4$\times$ & 17.1s \\
                      & Browser-Use & 4.6s & 32.6s & 7.1$\times$ & 21.2s \\
\addlinespace
Gemini 3.1 Flash-Lite & AX-tree     & 4.1s & 14.5s & 3.6$\times$ & 11.2s \\
                      & SoM         & 3.8s & 14.4s & 3.8$\times$ & 11.6s \\
                      & Pixel       & 3.1s & 15.8s & 5.1$\times$ & 11.8s \\
                      & Browser-Use & 4.5s & 23.1s & 5.1$\times$ & 17.7s \\
\addlinespace
GPT-5.4               & AX-tree     & 4.0s & 31.9s & 7.9$\times$ & 19.2s \\
                      & SoM         & 3.7s & 35.0s & 9.4$\times$ & 22.3s \\
                      & Pixel       & 4.2s & 23.0s & 5.5$\times$ & 14.2s \\
                      & Browser-Use & 4.6s & 36.2s & 7.8$\times$ & 24.3s \\
\addlinespace
GPT-5 mini            & AX-tree     & 4.5s & 32.1s & 7.2$\times$ & 18.9s \\
                      & SoM         & 4.3s & 32.7s & 7.7$\times$ & 19.9s \\
                      & Pixel       & 3.3s & 71.8s & 21.5$\times$ & 41.4s \\
                      & Browser-Use & 4.6s & 36.3s & 8.0$\times$ & 24.2s \\
\addlinespace
GPT-5.4 mini          & AX-tree     & 3.9s & 14.7s & 3.8$\times$ & 11.4s \\
                      & SoM         & 3.9s & 14.2s & 3.7$\times$ & 11.4s \\
                      & Pixel       & 3.6s & 14.2s & 3.9$\times$ & 10.7s \\
                      & Browser-Use & 4.2s & 17.6s & 4.2$\times$ & 13.7s \\
\addlinespace
Qwen3-VL-235B         & AX-tree     & 3.9s & 21.6s & 5.6$\times$ & 13.8s \\
                      & SoM         & 3.4s & 24.4s & 7.1$\times$ & 14.4s \\
                      & Pixel       & 2.9s & 26.8s & 9.3$\times$ & 16.0s \\
                      & Browser-Use & 4.1s & 45.0s & 10.9$\times$ & 28.1s \\
\addlinespace
UI-TARS-1.5-7B        & Native Pixel & 2.4s & 15.5s & 6.5$\times$ & 10.1s \\
\bottomrule
\end{tabular}
\caption{Full time efficiency on successful ComponentBench-Full tasks. Human and Agent columns report mean wall-clock duration over the same task set: the tasks each model--mode configuration solved (hence the Human column varies by row). Ratio is agent-to-human mean time.}
\label{tab:time-full}
\end{table}

\section{Representation trade-off tables}
\label{app:repr}

\begin{table}[h]
\centering
\small
\begin{tabular}{@{}lrrr@{}}
\toprule
\textbf{Model} & \textbf{SoM} & \textbf{Pixel} & \textbf{$\Delta$ (SoM$-$Pixel)} \\
\midrule
GPT-5 mini            & 78.5 & 48.9 & $+$29.6 \\
Gemini 3.1 Flash-Lite & 73.5 & 63.3 & $+$10.2 \\
Qwen3-VL-235B         & 54.4 & 50.5 & $+$3.9 \\
Gemini 3 Flash        & 87.1 & 85.4 & $+$1.7 \\
GPT-5.4 mini          & 74.7 & 77.1 & $-$2.4 \\
GPT-5.4               & 77.0 & 83.8 & $-$6.8 \\
\bottomrule
\end{tabular}
\caption{SoM--Pixel delta on ComponentBench-Full (\%). The SoM advantage is model-dependent, ranging from $+$29.6\% to $-$6.8\%.}
\label{tab:som-pixel-delta}
\end{table}

\begin{table}[h]
\centering
\small
\begin{tabular}{@{}lrrr@{}}
\toprule
\textbf{Model} & \textbf{Browser-Use} & \textbf{Mean(AX,SoM,Pix)} & \textbf{$\Delta$} \\
\midrule
Qwen3-VL-235B         & 78.8 & 60.6 & $+$18.2 \\
GPT-5 mini            & 87.0 & 70.2 & $+$16.8 \\
Gemini 3.1 Flash-Lite & 87.4 & 71.5 & $+$15.9 \\
GPT-5.4               & 90.4 & 80.8 & $+$9.6 \\
GPT-5.4 mini          & 85.8 & 77.0 & $+$8.8 \\
Gemini 3 Flash        & 95.2 & 87.4 & $+$7.8 \\
\bottomrule
\end{tabular}
\caption{Browser-Use advantage over mean non-Browser-Use pass rate (\%). The advantage is largest for weaker models.}
\label{tab:browser-use-gap}
\end{table}

\section{Additional analysis figures}
\label{app:additional-figures}
\begin{figure*}[p]
\centering
\includegraphics[width=\textwidth]{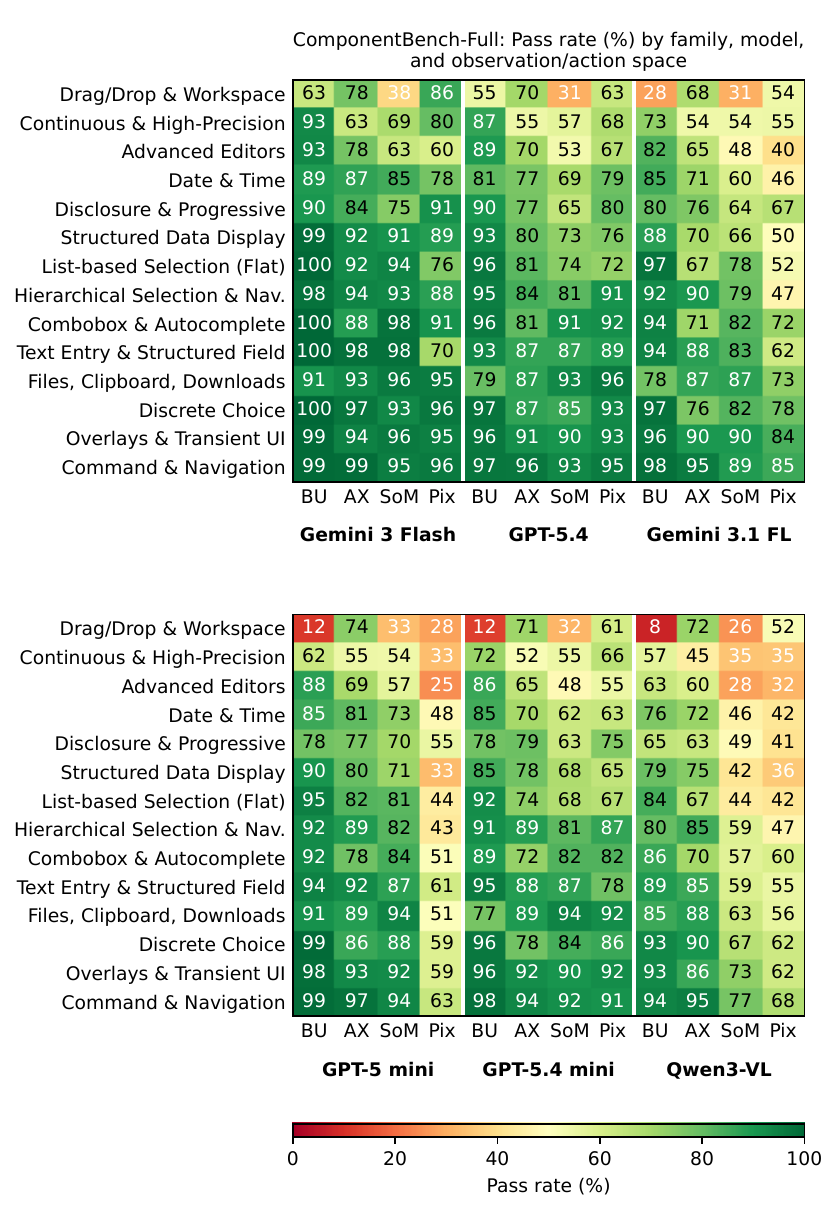}
\caption{Pass rate (\%) by component family, model, and observation mode on ComponentBench-Full. Families sorted by difficulty (hardest at top). The heatmap reveals where mode ordering inverts: Drag/Drop favors Pixel over Browser-Use; Advanced Editors strongly favor Browser-Use.}
\label{fig:family-heatmap}
\end{figure*}

\begin{figure}[h]
\centering
\includegraphics[width=\columnwidth]{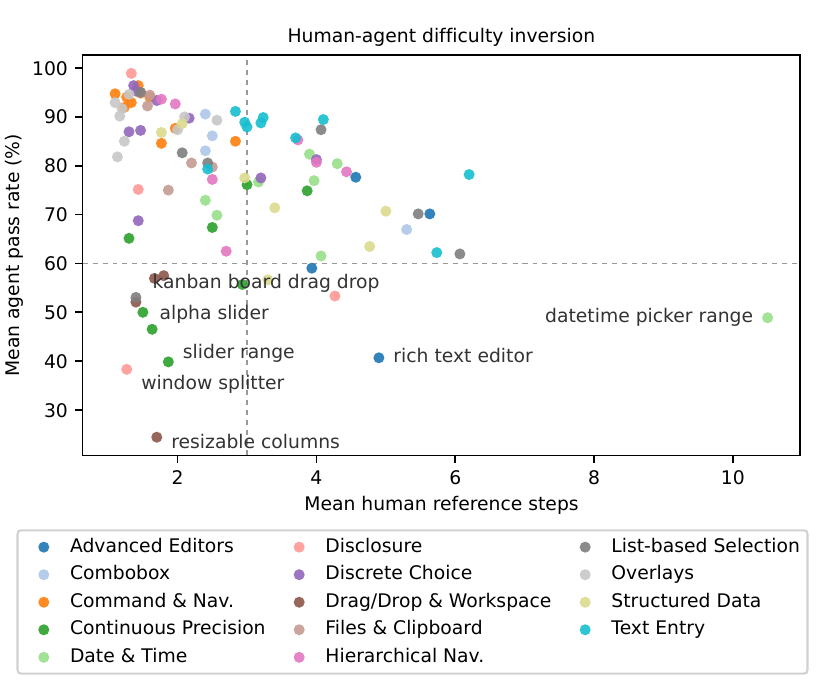}
\caption{Human-agent difficulty inversion. Each point is one canonical type. The lower-left quadrant contains components trivial for humans ($\leq$2 steps) but hard for agents ($<$60\% pass).}
\label{fig:inversion}
\end{figure}

\begin{figure}[h]
\centering
\includegraphics[width=\columnwidth]{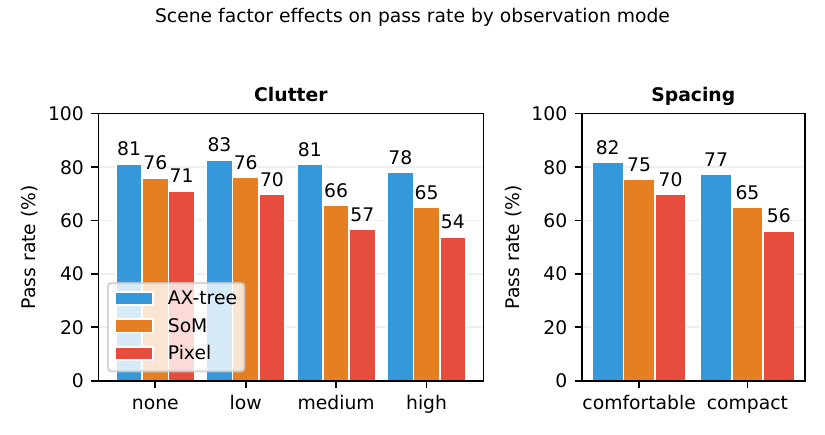}
\caption{Clutter and spacing disproportionately hurt Pixel mode, widening the AX-tree--Pixel gap from 10\% to 24\% (clutter) and 12\% to 21\% (spacing).}
\label{fig:scene-factors}
\end{figure}

\begin{figure}[h]
\centering
\includegraphics[width=\columnwidth]{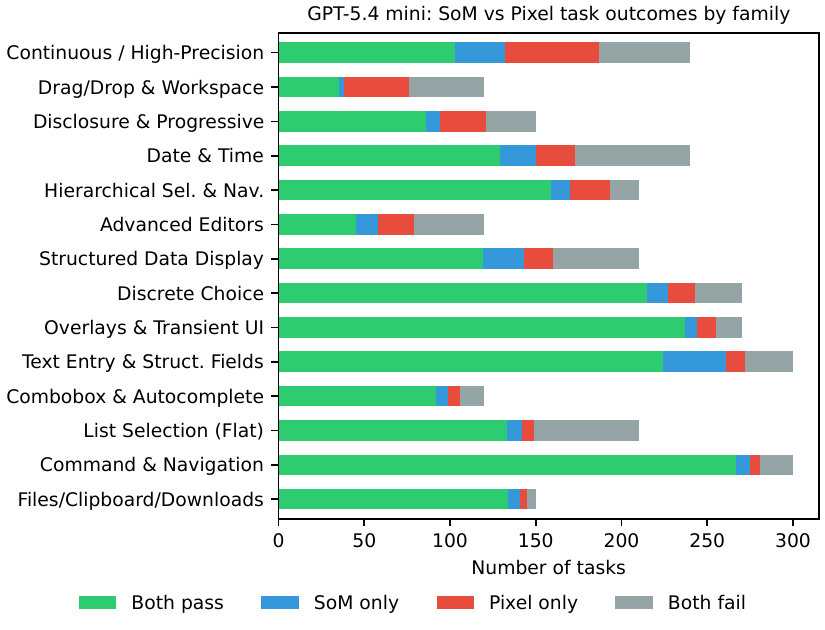}
\caption{GPT-5.4~mini SoM vs.\ Pixel task outcomes by component family. For each family, tasks are classified as both-pass, SoM-only, Pixel-only, or both-fail. Drag/Drop and Continuous Precision families show the strongest Pixel-over-SoM advantage.}
\label{fig:som-pixel-family}
\end{figure}

\begin{figure}[h]
\centering
\includegraphics[width=\columnwidth]{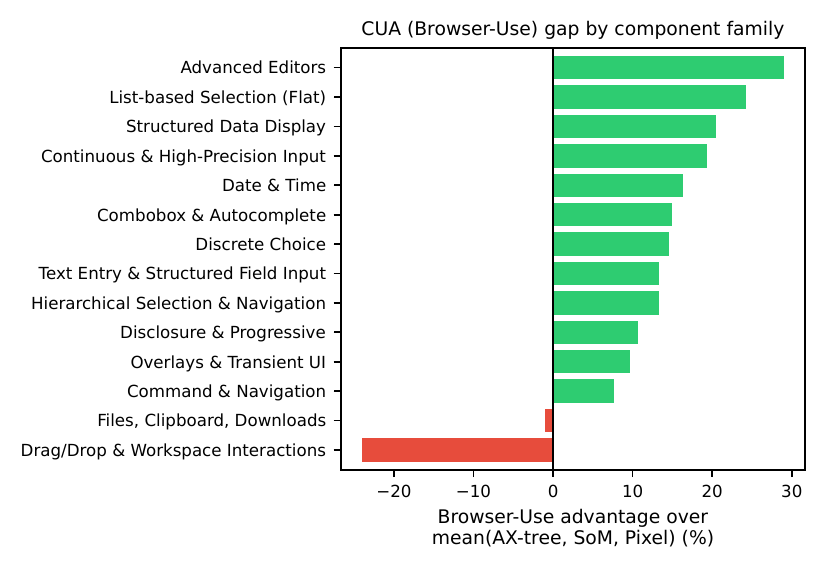}
\caption{Browser-Use (CUA) advantage by component family, averaged across the six models evaluated in all four regimes. The advantage ranges from $+$29\% (Advanced Editors) to $-$24\% (Drag/Drop), confirming that DOM-level tool access is not uniformly beneficial.}
\label{fig:cua-gap-family}
\end{figure}

\begin{figure}[h]
\centering
\includegraphics[width=\columnwidth]{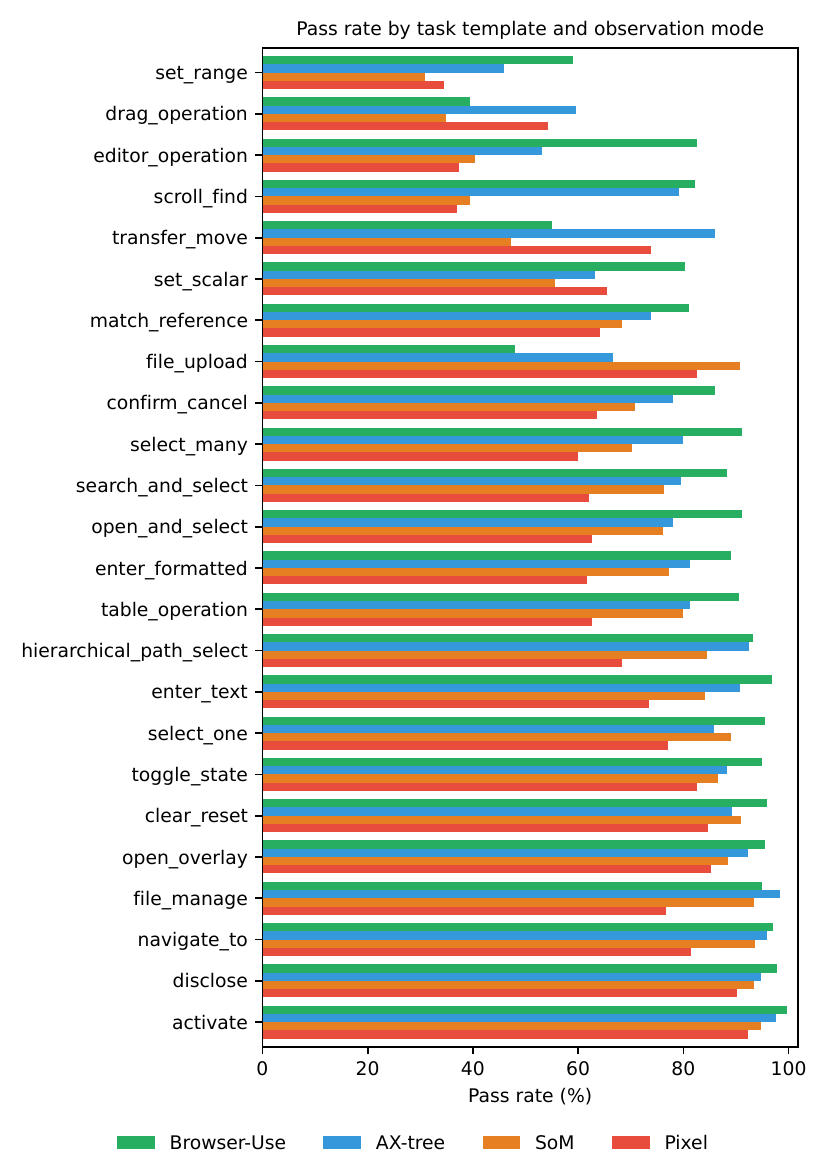}
\caption{Pass rate by task template and observation mode. Templates requiring spatial control (set\_range, drag\_operation) or complex editing (editor\_operation) are substantially harder than simple activation or disclosure tasks.}
\label{fig:template-passrate}
\end{figure}

\begin{figure}[h]
\centering
\includegraphics[width=\columnwidth]{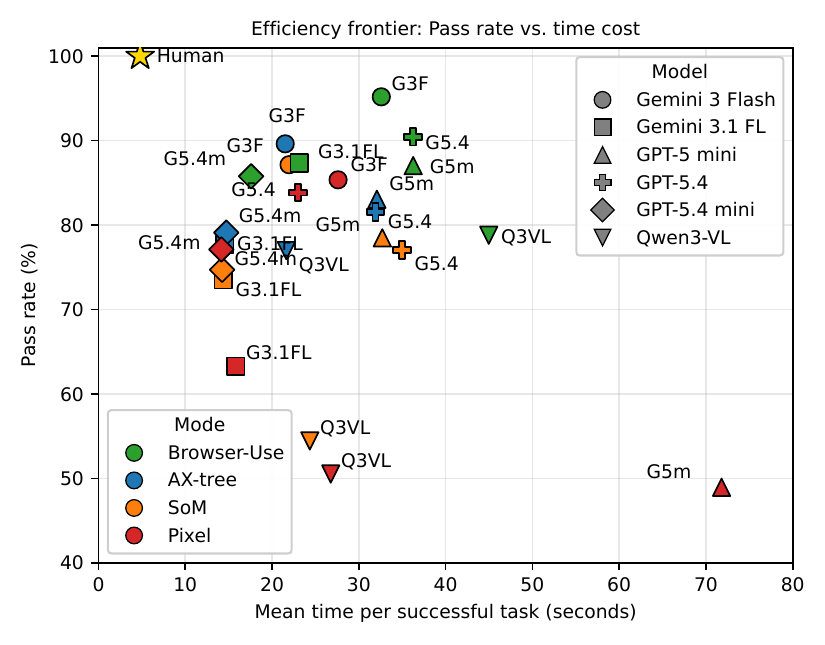}
\caption{Efficiency frontier: pass rate vs.\ mean time per successful task. Each point is one model-mode combination among the six models evaluated in the shared and Browser-Use regimes; the native UI-TARS configuration is omitted. The human reference point (100\% pass, 4.8s) is shown for comparison.}
\label{fig:efficiency-frontier}
\end{figure}

\begin{figure}[h]
\centering
\includegraphics[width=\columnwidth]{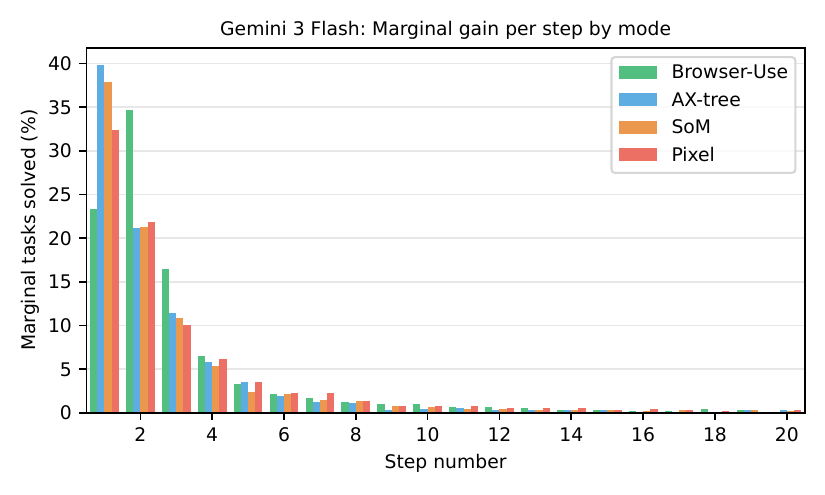}
\caption{Marginal tasks solved per additional step (Gemini~3~Flash). Most diagnostic value is concentrated in the first 5 steps; steps 6--20 contribute diminishing returns.}
\label{fig:marginal-steps}
\end{figure}

Figures~\ref{fig:family-heatmap}--\ref{fig:marginal-steps} provide the full per-model family heatmap, difficulty-inversion scatter, scene-factor effects, and per-template, efficiency, and step-curve breakdowns referenced in the main text.

\section{Difficulty axis validation}
\label{app:difficulty-validation}

\begin{table}[h]
\centering
\small
\setlength{\tabcolsep}{3pt}
\begin{tabular}{@{}lrrrr@{}}
\toprule
\textbf{Difficulty axis} & \textbf{Overall} & \textbf{AX-tree} & \textbf{Pixel} & \textbf{Browser-Use} \\
\midrule
Precision requirement       & $+$0.44 & $+$0.30 & $+$0.41 & $+$0.33 \\
Target acquisition          & $+$0.31 & $+$0.17 & $+$0.36 & $+$0.19 \\
Density / choice interf.    & $+$0.23 & $+$0.10 & $+$0.31 & $+$0.12 \\
Feedback dynamics           & $+$0.23 & $+$0.12 & $+$0.29 & $+$0.13 \\
Depth / layering            & $+$0.20 & $+$0.11 & $+$0.29 & $+$0.05 \\
Semantic observability      & $+$0.13 & $+$0.08 & $+$0.17 & $+$0.05 \\
Disambiguation load         & $+$0.09 & $+$0.04 & $+$0.15 & $+$0.02 \\
\bottomrule
\end{tabular}
\caption{Pearson correlation between intended difficulty axes and agent failure rate. Precision requirement is the most predictive axis overall; all axes predict Pixel failure more strongly than Browser-Use failure.}
\label{tab:axis-correlation}
\end{table}

\begin{table}[h]
\centering
\small
\begin{tabular}{@{}lrrrr@{}}
\toprule
\textbf{Tier} & \textbf{AX-tree} & \textbf{SoM} & \textbf{Pixel} & \textbf{Browser-Use} \\
\midrule
L0 (easy)   & 87.7\% & 85.9\% & 83.0\% & 92.8\% \\
L1 (medium) & 82.0\% & 75.1\% & 69.8\% & 87.6\% \\
L2 (hard)   & 77.4\% & 66.4\% & 57.6\% & 84.3\% \\
L3 (hard+)  & 71.4\% & 59.4\% & 49.5\% & 80.4\% \\
\bottomrule
\end{tabular}
\caption{Pass rate by intended difficulty tier and observation mode, averaged across all six models (excluding UI-TARS). The AX-tree--Pixel gap widens monotonically from 4.6\% (L0) to 21.9\% (L3).}
\label{tab:tier-validation}
\end{table}

\section{Ontology: 97 canonical component types}
\label{app:ontology}

The 14 families and their canonical types (with component count per family):

\begin{description}
\setlength{\itemsep}{1pt}
\item[Command \& Navigation (10):] button, icon\_button, link, menu\_button, split\_button, toolbar, breadcrumb, pagination, stepper, tabs
\item[Disclosure \& Progressive (5):] accordion, collapsible\_disclosure, carousel, feed\_infinite\_scroll, window\_splitter
\item[Text Entry \& Structured Field Input (10):] text\_input, textarea, password\_input, number\_input\_spinbutton, masked\_input, pin\_input\_otp, tags\_input, mentions\_input, search\_input, inline\_editable\_text
\item[Discrete Choice (9):] checkbox, checkbox\_group, checkbox\_tristate, radio\_group, switch, toggle\_button, toggle\_button\_group\_multi, segmented\_control, rating
\item[List-based Selection (Flat) (7):] select\_native, select\_custom\_single, select\_custom\_multi, select\_with\_search, listbox\_single, listbox\_multi, transfer\_list
\item[Combobox \& Autocomplete (4):] combobox\_editable\_single, combobox\_editable\_multi, autocomplete\_freeform, autocomplete\_restricted
\item[Hierarchical Selection \& Navigation (7):] menu, menubar, context\_menu, tree\_view, tree\_select, tree\_grid, cascader
\item[Continuous \& High-Precision Input (8):] slider\_single, slider\_range, alpha\_slider, meter, progress\_bar, color\_picker\_2d, color\_swatch\_picker, color\_text\_input
\item[Date \& Time (8):] date\_picker\_single, date\_picker\_range, datetime\_picker\_single, datetime\_picker\_range, time\_picker, calendar\_embedded, date\_input\_text, time\_input\_text
\item[Overlays \& Transient UI (9):] dialog\_modal, drawer, popover, tooltip, hover\_card, toast\_snackbar, alert\_dialog\_confirm, notification\_center, tour\_teaching\_tip
\item[Structured Data Display (7):] table\_static, data\_table\_sortable, data\_table\_paginated, data\_table\_filterable, data\_grid\_editable, data\_grid\_row\_selection, virtual\_list
\item[Files, Clipboard, Downloads (5):] file\_upload\_button, file\_dropzone, file\_list\_manager, clipboard\_copy, download\_trigger
\item[Drag/Drop \& Workspace Interactions (4):] drag\_drop\_sortable\_list, drag\_drop\_between\_lists, kanban\_board\_drag\_drop, resizable\_columns
\item[Advanced Editors (4):] rich\_text\_editor, markdown\_editor, code\_editor, json\_editor
\end{description}

\section{Hardest canonical component types}
\label{app:hardest}

\begin{table}[h]
\centering
\small
\setlength{\tabcolsep}{3pt}
\begin{tabular}{@{}lrrll@{}}
\toprule
\textbf{Component type} & \textbf{Agent \%} & \textbf{Human steps} & \textbf{Family} \\
\midrule
resizable\_columns        & 24.4 & 1.7 & Drag/Drop \\
window\_splitter           & 38.3 & 1.3 & Disclosure \\
slider\_range              & 39.9 & 1.9 & Continuous \\
rich\_text\_editor          & 40.7 & 4.9 & Adv.\ Editors \\
meter                     & 46.5 & 1.6 & Continuous \\
datetime\_picker\_range     & 48.9 & 10.5 & Date/Time \\
alpha\_slider              & 50.0 & 1.5 & Continuous \\
kanban\_board\_drag\_drop    & 52.1 & 1.4 & Drag/Drop \\
select\_native             & 53.1 & 1.4 & List Selection \\
feed\_infinite\_scroll      & 53.3 & 4.3 & Disclosure \\
color\_picker\_2d           & 55.7 & 2.9 & Continuous \\
virtual\_list              & 56.7 & 3.3 & Structured Data \\
drag\_drop\_between\_lists   & 56.9 & 1.7 & Drag/Drop \\
drag\_drop\_sortable\_list   & 57.5 & 1.8 & Drag/Drop \\
code\_editor               & 59.0 & 3.9 & Adv.\ Editors \\
\bottomrule
\end{tabular}
\caption{The 15 hardest canonical types by mean agent pass rate (averaged across all models and modes, excluding UI-TARS). Human steps column shows the mean normalized human reference steps. Components with $\leq$2 human steps but $<$60\% agent pass rate represent the human-agent difficulty inversion.}
\label{tab:hardest-types}
\end{table}

\section{Failure taxonomy details and case studies}
\label{app:failure-analysis}

\subsection{Method}

The deterministic labeling in Table~\ref{tab:failure-taxonomy} parses every failed Pixel/SoM/AX-tree trace (8,864 traces across Gemini~3~Flash, Gemini~3.1~Flash-Lite, GPT-5~mini, GPT-5.4, and GPT-5.4~mini) into its action sequence and extracts feature flags: whether a drag was emitted, whether the target control was ever interacted with, whether coordinates repeat across steps, and whether a value was typed. Each trace starts from a component-driven category prior (e.g., slider tasks default toward continuous calibration) and is refined or overridden by the trace evidence; \texttt{repetition\_or\_no\_progress\_loop} is retained as a residual category only when the trace does not support a more specific mechanism, with \texttt{other\_or\_unclear} as the final fallback. The Layer-2 column maps each failed run's LLM-assigned \texttt{primary\_failure\_family} and secondary tags onto the same nine categories; these labels are produced by GPT-5.4 (high reasoning effort) reading each failed run's full action log and screenshots under a fixed JSON output schema. Categories that require low-level pointer evidence (continuous calibration, drag execution) are not separable in the Layer-2 labels and are folded into their nearest semantic categories there.

\subsection{Adversarially reviewed case studies}

We selected 20 representative failed traces spanning the major categories, wrote a mechanism narrative for each by reading the full action log and screenshots alongside the Layer-2 diagnosis, and then had an \emph{independent adversarial reviewer}---a separately prompted Claude Opus 4.8 \citep[\texttt{claude-opus-4-8};][]{anthropic2026opus48} pass instructed to re-open the same evidence with skepticism and refute each narrative---re-check every case. Each case received a single review pass, issued through the Claude Code agent harness at its default sampling and reasoning settings (i.e., not a deterministic decode); the review script ships with the benchmark. Fifteen cases were confirmed as labelled; five were relabelled by the reviewer, and we report the reviewer-adjusted category. Three abbreviated examples:

\paragraph{Continuous calibration (\texttt{meter-mui-T09}, GPT-5.4~mini, Pixel).} Instruction: drag the Server~B load meter to 42\% in a three-row table. The agent targets the correct row from the first step and successfully drags the bar (15\%$\to$28\% across the run), but the bar spans only $\sim$100px for 0--100\%, so each coarse drag overshoots or undershoots; the agent issues progressively smaller nudges and times out at $\sim$28\%. The failure is value calibration, not grounding or instance selection.

\paragraph{Target acquisition in drag (\texttt{drag\_drop\_between\_lists-antd-T01}, GPT-5~mini, Pixel).} Instruction: drag \emph{Editor} from Available to Assigned roles. The drag primitive works---items visibly move---but the very first drag grabs the row \emph{above} the intended one (aimed at $y{=}350$ for Editor; that $y$ is the Admin row), placing Admin into Assigned. The agent later also moves Editor but never removes the mis-dragged Admin, so the committed set fails exact-set verification. An off-by-one-row grab, not a drag-execution failure.

\paragraph{Perception error masquerading as a loop (\texttt{kanban\_board\_drag\_drop-antd-T04}, GPT-5.4~mini, Pixel).} Instruction: reorder the Review column to match a reference panel. The agent misreads the reference (quoting the column's own current order back as the target), concludes the board already matches, and spends all 20 steps re-asserting completion without ever issuing a drag. The root cause is a wrong belief about the target state; the loop is only the symptom.

The five reviewer adjustments are themselves instructive: in each, a plausible category (e.g., ``calibration error'') was overturned by finer trace reading (e.g., the agent was monotonically ratcheting a spinner in the wrong direction---a no-progress loop, not calibration). All 20 case studies with trace pointers are released alongside the benchmark.

\section{Human reference validation with additional annotators}
\label{app:human-validation}

The human reference traces in the main experiments come from two passes by a single annotator (Section~\ref{sec:construction-human}). To test whether the efficiency comparisons depend on that annotator's idiosyncrasies, we collected recordings from \textbf{two additional annotators} on a fixed, pre-specified \textbf{278-task stratified validation subset}: one task per realized (canonical type, library) pair, chosen by a fixed-seed randomized search (seed 2026) that minimizes deviation from the global difficulty distribution before inspecting any new results. The subset covers all 97 canonical types and all 14 families (bucket distribution easy 93 / mid 94 / hard 91), and contains 278 rather than $97 \times 3 = 291$ tasks because not every canonical type is implemented in every library.

Each additional annotator followed the same protocol as the original reference: two recorded passes per task with the shorter successful pass kept, cleaned with the same normalization pipeline. All three annotators completed all 278 tasks successfully. Table~\ref{tab:annotator-agreement} summarizes agreement on action tasks (258--265 per annotator; the remainder are hover-only tasks with zero normalized actions). Pairwise statistics are computed over the tasks where both annotators in the pair recorded at least one normalized action; hover-only tasks are excluded.

\begin{table}[h]
\centering
\small
\begin{tabular}{@{}lrr@{}}
\toprule
\textbf{Metric} & & \textbf{Value} \\
\midrule
Mean normalized steps (Original / A1 / A2) & & 2.92 / 2.90 / 3.14 \\
Median normalized steps (all annotators) & & 2 \\
Pairwise step-count Pearson correlation & & 0.79--0.94 \\
Exact step-count agreement & & 71\%--84\% \\
Mean absolute step-count difference & & 0.32--0.64 \\
\bottomrule
\end{tabular}
\caption{Agreement between the original reference annotator and two additional annotators (A1, A2) on the 278-task validation subset (action tasks only). Human-to-human variation is small at the aggregate level.}
\label{tab:annotator-agreement}
\end{table}

On a like-for-like per-task basis, the human-to-human action-count ratio is \textbf{1.05$\times$} (each annotator versus the median of the others), whereas agents on the same subset take \textbf{1.27--3.02$\times$} the human action count on tasks they solve (Appendix~\ref{app:stability}). The agent--human efficiency gap is therefore several times larger than inter-annotator variation. We accordingly present the reference traces as \emph{practical successful references} rather than claims of human optimality; the efficiency conclusions in Section~\ref{sec:step-efficiency} are robust to the choice of annotator.

The recording sessions also served as an independent task-quality spot check: annotators could flag broken, ambiguous, or miswired tasks during recording, and no task in the subset was flagged---evidence against pervasive implementation or specification defects in this stratified subset.

\section{Repeated-run stability}
\label{app:stability}

Each model--mode combination in the main tables was evaluated in a single deterministic run. To quantify run-to-run variation, we ran each selected model--mode cell \textbf{twice in total} on the same 278-task validation subset (Appendix~\ref{app:human-validation}) at an identical harness commit and model endpoint, for four models from two providers (Gemini~3~Flash, Gemini~3.1~Flash-Lite, GPT-5~mini, GPT-5.4~mini) under the two regimes that stress different variance sources: pure-visual Pixel and tool-rich Browser-Use. We report nonparametric bootstrap 95\% confidence intervals (resampling tasks, $B{=}10{,}000$, seed 2026).

\begin{table}[h]
\centering
\small
\setlength{\tabcolsep}{4pt}
\begin{tabular}{@{}llrrrrr@{}}
\toprule
\textbf{Model} & \textbf{Mode} & \textbf{Mean pass (\%)} & \textbf{Max dev} & \textbf{Agree\%} & \textbf{Jaccard} & \textbf{95\% CI} \\
\midrule
Gemini 3 Flash        & Browser-Use & 95.1 & 1.1 & 97.5 & 0.974 & [92.6, 97.3] \\
Gemini 3 Flash        & Pixel       & 87.4 & 0.7 & 90.6 & 0.898 & [84.0, 90.8] \\
Gemini 3.1 Flash-Lite & Browser-Use & 87.1 & 0.7 & 94.2 & 0.936 & [83.3, 90.6] \\
Gemini 3.1 Flash-Lite & Pixel       & 77.0 & 0.0 & 86.3 & 0.837 & [72.5, 81.3] \\
GPT-5 mini            & Browser-Use & 86.7 & 1.4 & 92.1 & 0.913 & [82.9, 90.3] \\
GPT-5 mini            & Pixel       & 52.5 & 0.0 & 82.7 & 0.718 & [47.1, 57.9] \\
GPT-5.4 mini          & Browser-Use & 85.4 & 1.1 & 91.0 & 0.900 & [81.5, 89.0] \\
GPT-5.4 mini          & Pixel       & 80.6 & 1.4 & 87.8 & 0.859 & [76.3, 84.7] \\
\bottomrule
\end{tabular}
\caption{Repeated-run stability on the 278-task subset (2 runs per cell, same harness commit and endpoint). Max dev: run-to-run pass-rate deviation (\%). Agree\%: task-level exact pass/fail agreement. Jaccard: overlap of passed-task sets. CI: bootstrap 95\% interval on the mean pass rate.}
\label{tab:stability}
\end{table}

Three observations. \textbf{(1) Aggregate pass rates are stable}: run-to-run deviation is at most 1.4\% (median 0.9\%), far smaller than the mode and model effects in the main results, and the headline orderings (Gemini~3~Flash strongest; Browser-Use $>$ Pixel for every model; GPT-5~mini Pixel weakest) are identical in both runs. \textbf{(2) Aggregate stability can hide task-level churn}: GPT-5~mini Pixel has an identical pass rate in both runs (52.5\%) yet 17.3\% of tasks flip outcome, with pass$\to$fail and fail$\to$pass flips canceling in the aggregate. We therefore report task-level agreement and Jaccard alongside pass rates; conclusions about individual borderline tasks should be read with this churn in mind. Flips concentrate on mid/hard tasks and on the drag/slider components already identified as brittle. \textbf{(3) Interaction counts are far more stable than wall-clock time}: mean successful-task action counts change by at most 0.40 steps between runs, while mean durations shift by up to 11 seconds---so we base efficiency claims on action counts rather than latency-confounded durations. On the same subset, the per-task action-count ratio of each successful agent run to the human median ranges from 1.27$\times$ (GPT-5.4~mini and Gemini~3.1~Flash-Lite, Pixel) to 3.02$\times$ (GPT-5~mini, Pixel), the like-for-like comparison quoted against the 1.05$\times$ human--human baseline in Appendix~\ref{app:human-validation}.

\section{Experimental setup details}
\label{app:setup}

This section documents the model access, coordinate handling, and framework configuration used in our experiments.

\subsection{BrowserGym-based evaluation (AX-tree, SoM, Pixel)}

All models except UI-TARS-1.5-7B and Opus~4.6 are evaluated through a shared BrowserGym-based harness \citep{dechezelles2024browsergym} with a \textbf{1280$\times$720} viewport, a maximum of \textbf{20 steps} per task, and a \textbf{600-second} per-task wall-clock budget; the BrowserGym modes additionally enforce a 300-second per-step timeout. The three observation/action spaces (AX-tree, SoM, Pixel) differ only in what the agent sees and how it refers to targets; the underlying page, verifier, and termination logic are identical.

In Pixel mode, different model families use different coordinate conventions:

\begin{itemize}
\setlength{\itemsep}{2pt}
\item \textbf{Gemini~3~Flash and Gemini~3.1~Flash-Lite} are accessed via the \textbf{Google~AI~Studio API}. These models output coordinates in a normalized 0--1000 space; the harness linearly maps them to screen pixels.
\item \textbf{Qwen3-VL-235B-FP8} is served locally via \textbf{vLLM} (FP8 quantization, tensor-parallel). It also outputs 0--1000 normalized coordinates, mapped to pixels by the harness.
\item \textbf{GPT-5.4} and \textbf{GPT-5.4~mini} are accessed via the \textbf{OpenAI API} directly. These models output raw pixel coordinates; no coordinate transformation is applied.
\item \textbf{GPT-5~mini} is accessed via a \textbf{Duke LiteLLM} proxy (OpenAI-compatible endpoint). It also outputs raw pixel coordinates with no transformation.
\end{itemize}

All models receive a screenshot as a base64-encoded image. In AX-tree mode, the accessibility-tree text is appended to the prompt. In SoM mode, numbered bounding-box overlays are rendered onto the screenshot.

\subsection{UI-TARS-1.5-7B (native pixel mode)}

UI-TARS-1.5-7B \citep{qin2025uitars} is evaluated through its own native harness rather than the shared BrowserGym pipeline, to match its training-time interface. The model is served via \textbf{vLLM} (bfloat16, single GPU, 32K context window, \texttt{gpu\_memory\_utilization=0.90}). Screenshots are rescaled using \texttt{smart\_resize} (from the Qwen-VL codebase) before being sent to the model. The model predicts coordinates in the resized-image space; the agent maps them back to the original 1280$\times$720 screen pixels. The conversation uses a \textbf{multi-turn} format with a 4-turn sliding window for history, matching the model's training setup. Temperature is set to 0.0, with up to 3 retry attempts per step for invalid actions.

\subsection{Opus~4.6 (Core Pixel only)}

Opus~4.6 is evaluated only on ComponentBench-Core (912 tasks) in Pixel mode. Screenshots are captured at the native 1280$\times$720 viewport and \textbf{anisotropically resized to 1024$\times$768} (16:9 to 4:3, no padding or cropping) before being sent to Claude via Anthropic's \textbf{computer-use tool} interface \citep{anthropic2026computeruse}. Claude outputs coordinates in 1024$\times$768 space; the agent maps them back to 1280$\times$720 using separate horizontal and vertical scale factors. The target resolution follows Anthropic's XGA recommendation for optimal model accuracy.

\subsection{Browser-Use mode}

The Browser-Use observation/action space uses the \texttt{browser-use} framework, which provides the agent with screenshot-based interaction together with serialized DOM information and grounded element references. Key parameters:

\begin{itemize}
\setlength{\itemsep}{2pt}
\item Viewport: 1280$\times$720, headless Chromium
\item \texttt{use\_vision=True}, \texttt{use\_thinking=True}, \texttt{flash\_mode=False}
\item \texttt{max\_actions\_per\_step=4}, \texttt{max\_failures=3}, \texttt{step\_timeout=120s}
\item \texttt{highlight\_elements=False} (no visual element highlighting)
\item Maximum 20 steps per task; the initial page-load step is excluded from step-count analysis
\end{itemize}

Gemini models are accessed via the Google~AI~Studio API; GPT models are accessed via the OpenAI API or Duke LiteLLM proxy. Qwen3-VL-235B's Browser-Use runs are served through Amazon Bedrock's OpenAI-compatible endpoint (\texttt{qwen.qwen3-vl-235b-a22b}); Bedrock does not document its serving precision, so these runs may differ slightly from the locally served FP8 configuration used for its AX-tree, SoM, and Pixel runs.

\end{document}